\pdfoutput=1

\documentclass[11pt]{article}

\usepackage[final]{acl}

\usepackage{times}
\usepackage{latexsym}
\usepackage{amsmath}
\usepackage{amssymb}
\usepackage{mathtools}
\usepackage{amsthm}
\usepackage{bbm}
\usepackage{enumitem}
\usepackage{graphicx}
\usepackage{capt-of}
\usepackage{varwidth}
\usepackage[capitalize,noabbrev]{cleveref}
\usepackage{booktabs} 

\usepackage{amsfonts} 
\usepackage{multirow}
\usepackage{arydshln}
\usepackage{colortbl}  
\usepackage{xcolor}    
\usepackage{tcolorbox}
\usepackage{tikz}

\definecolor{darkgray}{rgb}{0.4, 0.4, 0.4}
\definecolor{ddarkgray}{rgb}{0.2, 0.2, 0.2}
\definecolor{darkgreen}{rgb}{0,100,0}
\newcommand{\db}[1]{\textcolor{darkblue}{\bf \selectfont \,(#1)}}

\newcommand{\defeq}[0]{\mathrel{\stackrel{\textnormal{\tiny def}}{=}}}
\newcommand{\code}[1]{{\ttfamily#1}}

\definecolor{darkblue}{rgb}{0.0,0.0,0.5}
\definecolor{purple}{rgb}{0.5,0.0,0.5}

\crefname{theorem}{Theorem}{}
\crefname{hypothesis}{Hyp.}{Hypotheses}
\crefname{assumption}{Assumption}{}
\crefname{prop}{Proposition}{}

\theoremstyle{plain}
\newtheorem{theorem}{Theorem}[section]

\newtheorem{proposition}[theorem]{Proposition}

\usepackage{subcaption}
\usepackage{threeparttable}
\usepackage[T1]{fontenc}

\usepackage[utf8]{inputenc}

\usepackage{microtype}

\usepackage{inconsolata}

\title{Exploiting Contextual Knowledge in LLMs through $\mathcal{V}$-usable \\Information based Layer Enhancement }

\author{
Xiaowei Yuan$^{1,2,3}$, Zhao Yang$^{4}$, Ziyang Huang$^{1,2}$, Yequan Wang$^{3}$, Siqi Fan$^{5}$, \\\textbf{Yiming Ju}$^{3}$, \textbf{Jun Zhao}$^{1,2}$, \textbf{Kang Liu}$^{1,2}$\\
$^1$The Key Laboratory of Cognition and Decision Intelligence for Complex Systems,\\
Institute of Automation, Chinese Academy of Sciences\\
$^2$School of Artificial Intelligence, University of Chinese Academy of Sciences\\
$^3$Beijing Academy of Artificial Intelligence
$^4$Meituan\\
$^5$University of Electronic Science and Technology of China \\
}

\begin{document}
\maketitle
\begin{abstract}
Large Language Models (LLMs) have demonstrated remarkable capabilities in various tasks, yet they often struggle with context-faithfulness generations that properly reflect contextual knowledge. While existing approaches focus on enhancing the decoding strategies, they ignore the fundamental mechanism of how contextual information is processed within LLMs’ internal states. As a result, LLMs remain limited in their ability to fully leverage contextual knowledge. In this paper, we propose Context-aware Layer Enhancement (CaLE), a novel intervention method that enhances the utilization of contextual knowledge within LLMs' internal representations. By employing $\mathcal{V}$-usable information analysis, CaLE strategically amplifies the growth of contextual information at an optimal layer, thereby enriching representations in the final layer. Our experiments demonstrate that CaLE effectively improves context-faithful generation in Question-Answering tasks, particularly in scenarios involving unknown or conflicting contextual knowledge.
\end{abstract}

\section{Introduction}
\begin{figure}[t]
    \centering
    \includegraphics[width=0.85\linewidth]{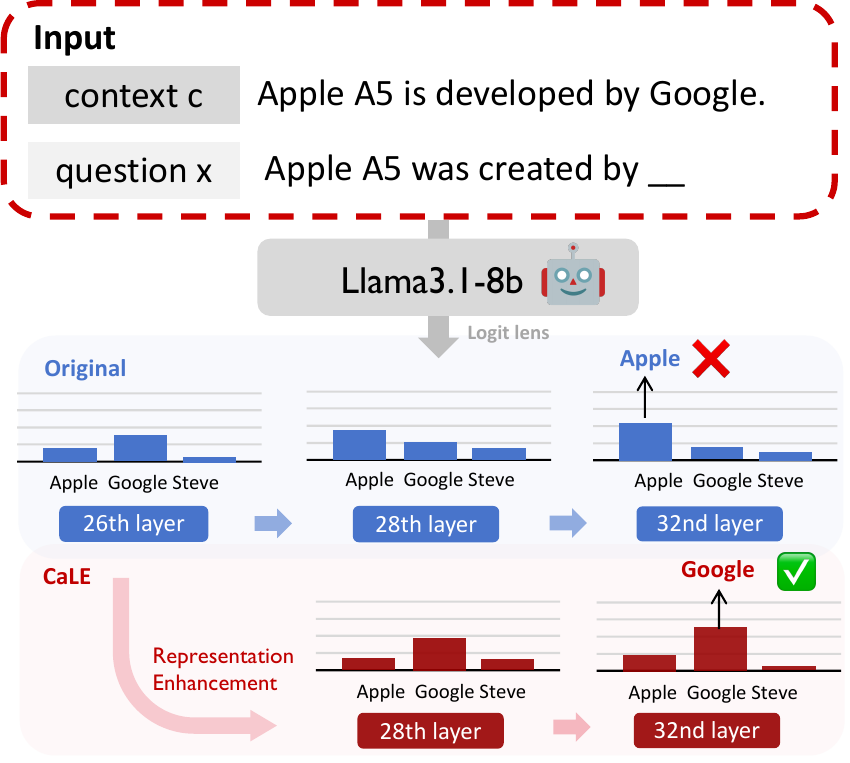}
    \caption{An illustration of CaLE Method.}
    \label{fig:model}
\end{figure}

Large Language Models (LLMs) have demonstrated remarkable capabilities in various tasks, yet they face significant challenges, including hallucination and outdated knowledge~\cite{journals/csur/JiLFYSXIBMF23,zhao2024}. Retrieval-Augmented Generation (RAG) has emerged as a promising approach to address these limitations by incorporating external knowledge sources into the generation process~\cite{ram-etal-2023-context,gao2024}. The concept of context-faithfulness—the ability to generate responses that accurately reflect provided contextual information—has thus become crucial for LLM applications~\cite{conf/emnlp/ZhouZPC23, ShiHLTZY24}. Nevertheless, these models often struggle to properly utilize external contextual information, particularly when it conflicts with their pre-existing parametric knowledge~\cite{Xie0CL024}.
As illustrated in the upper part of Figure~\ref{fig:model}, despite the presence of context indicating "Google", the model still generates a unfaithful output "Apple".

Existing efforts~\cite{ShiHLTZY24, JinCY0XLJ0024} to enhance the context-faithfulness of LLMs primarily focus on modifying decoding strategies or reweighting knowledge-aware neurons. 
The optimized decoding strategies~\cite{ShiHLTZY24,CLeHE,YuanYWLZL24} focus on the contrastive mechanism~\cite{LiHFLEHZL23} to ensure a greater reliance on external information.
Another line of research is to explore the internal neurons within models~\cite{IRCAN}. They aim to identify and reweight the neurons that are crucial in processing contextual cues. However, these methods are only applicable to predefined data formats, such as triplet facts or multiple-choice questions, thereby limiting their effectiveness in complex scenarios~\cite{JinCY0XLJ0024}.

Recent studies~\cite{abs-2304-13734, chen} have demonstrated that LLMs preserve the highly-concentrated information within their internal states. \citet{skean} further reveals that intermediate layers often yield more informative representations for downstream tasks than the final layer. These findings imply that, in RAG tasks, the contextual information within internal states may not always increase monotonically towards the output layer. As illustrated in Figure~\ref{fig:model}, the correct answer ("Google") attains the top probability at the 26th layer but not the final layer.
Therefore, we propose to explore the contextual information retained of LLMs' internal states for faithful generations.

We conduct a investigation on contextual information flow across model layers utilizing $\mathcal V$-usable information~\cite{XuZSSE20, EthayarajhCS22}. 
It measures the contribution that the inner states of model can help generate the contextual answer. Our findings reveal significant fluctuations in contextual information, which could lead to under-utilization of the given context. This fluctuation may disclose the inherent deficiency in processing the contextual information of current LLMs based on Transformer, and thus present a critical intervention point for preserving and enhancing contextual information flow, potentially improving the context-faithfulness of LLMs.

To remedy the above issues, this paper proposes a \underline{C}ontext-\underline{a}ware \underline{L}ayer \underline{E}nhancement (\text{CaLE}) method, which exploits contextual knowledge within model's internal representations from a layer-specific perspective. Based on $\mathcal{V}$-usable information, CaLE identifies the context-aware layer in either a supervised or unsupervised manner, which exhibits the highest contextual information. Then it enhances the layer representations through amplification or residual connections. As a result, the contextual information relevant to the target answers is effectively enriched.
As shown in Figure~\ref{fig:model}, CaLE identifies the 26-$th$ layer that encodes rich information about the correct answer ("Google") and enhances its representations, facilitating accurate response generation at the final layer. 

Experiments on CounterFact~\cite{MengBAB22}, Natural Questions (NQ)~\cite{NQ_KwiatkowskiPRCP19}, SQuAD~\cite{squad} and StrategyQA~\cite{strategyqa} datasets demonstrate that CaLE significantly improves context-faithful generation in downstream tasks. Furthermore, CaLE's enhancements to context utilization are independent of and complementary to various decoding strategies, enabling cumulative improvements in the faithfulnes of LLMs. This orthogonality to existing decoding methods underscores the versatility of our approach.

The contributions of this paper are as follows:
\begin{itemize}
    \item Through experimental analysis, we find that  LLMs often exhibit a characteristic information fluctuation across the intermediate layers, with certain layers maintaining a high increasing context-faithful information, followed by a plateau or decrease in the deeper layers.
    \item To mitigate the negative effective of the contextual information degradation, CaLE proposes a context-aware layer identification method to determine an optimal intervening position. Through amplification or residual connect, the further enhancement will lead to richer representations in the final layer.
\end{itemize}

\section{Information Flow Analysis based on \( \mathcal{V} \)-usable information} \label{sec:background}
First, we introduce a method for measuring the contribution of the inner states of the model to the faithfulness of its responses, specifically focusing on how to quantitatively analyze the contextual information contained within each layer’s state. Building on this, we analyze the flow of contextual information across different layers in various models, using the CounterFact~\cite{MengBAB22} dataset to examine the variations in the flow.

\subsection{$\mathcal{V}$-usable Information}
Unlike Shannon’s MI and in violation of the data processing inequality, \textit{$\mathcal{V}$-usable information} can be created through computation~\cite{XuZSSE20,EthayarajhCS22}.
It reflects the ease with which a model family $\mathcal{V}$ can predict the correct answer $\mathrm{Y}$ given specific input hidden states $h_l$ at layer $l$~\cite{Ju0DYRL24}.
\begin{equation} \label{eq:def_v}
    I_\mathcal V (\boldsymbol{h}_l \rightarrow Y) = H_\mathcal V(Y) - H_\mathcal V(Y|\boldsymbol{h}_l)
\end{equation}

\noindent where $H_\mathcal V(Y)$ and $H_\mathcal V(Y|h_l)$ denote the \textit{predictive $\mathcal V$-entropy} and the \textit{conditional $\mathcal V$-entropy}. The latter can be estimated through the following equations:
\begin{align}
    H_\mathcal V(Y) &= \underset{f \in \mathcal V}{\text{inf}} \mathbb E [-\ln f[\varnothing](Y)] \\
    H_\mathcal V(Y|\boldsymbol{h}_l) &= \underset{f \in \mathcal V}{\text{inf}} \mathbb E [-\ln f[\boldsymbol{h}_l](Y)]
\end{align}
where the function $f[\cdot]$ produces a probability distribution over the vocabulary. Put simply, we use the logit lens~\cite{nostalgebraist2020interpreting} with softmax function here.
\begin{equation}
    f[\boldsymbol{h}_{l}](Y) =\frac{e^{v_k }}{\sum_{j\in {|Vocab|}} e^{ v_j }}
\end{equation}
where $v = \text{LogitLens}(\boldsymbol{h}_l)$\footnote{Detailed formula can be found in Appendix~\ref{appendix:logitlens}.} represents the logit vector at layer $l$. The subscript $k$ represents the token index corresponding to $Y$.

To evaluate the variations across layers, we adopt $-H_\mathcal{V}$ for observation based on:
\begin{equation*}
    \Delta I_\mathcal V = I_\mathcal V (\boldsymbol{h}_l \rightarrow Y)-I_\mathcal V (\boldsymbol{h}_{l-1} \rightarrow Y) = \Delta -H_\mathcal{V}
\end{equation*}

\begin{figure}[t] 
    \centering
    \begin{subfigure}{0.23\textwidth}
        \includegraphics[width=\textwidth]{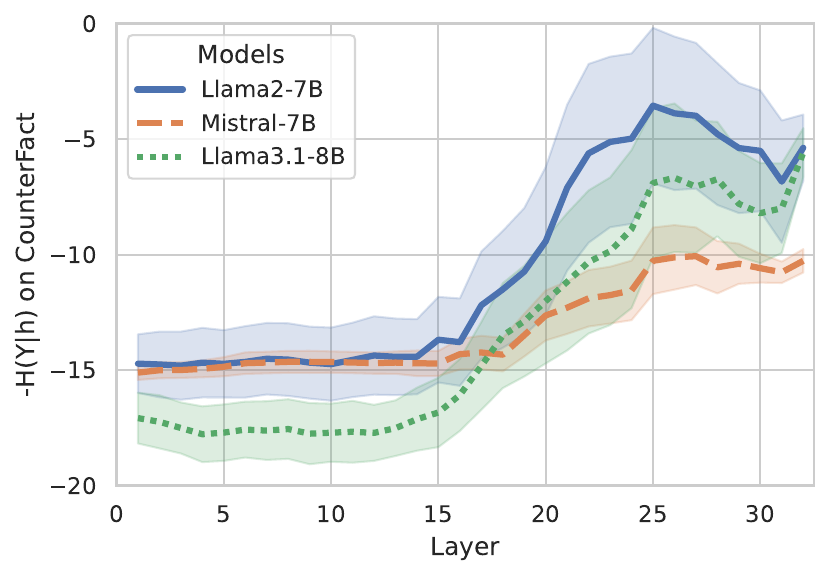}
        \caption{$-H_\mathcal V$ with models of similar sizes}
        \label{fig:entropy_cf}
    \end{subfigure}
\hfill
    \begin{subfigure}{0.23\textwidth}
        \includegraphics[width=\textwidth]{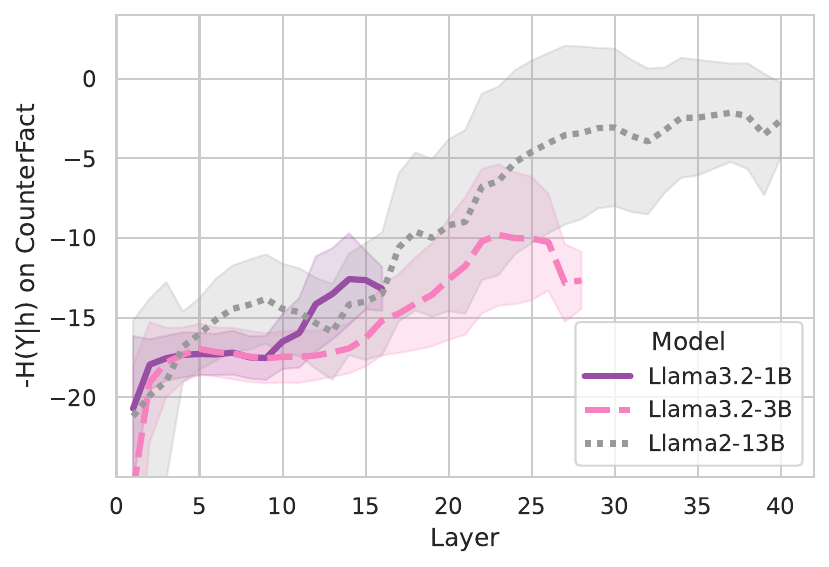}
        \caption{$-H_\mathcal V$ with models from the Llama series}
        \label{fig:entropy_nq}
    \end{subfigure}
    \caption{Visualization of Information Flow. The vertical axis represents the variation in $\mathcal{V}$-information, as reflected by the $-H_\mathcal{V}$ metric. The horizontal axis denotes the information content across different layers, while the shaded region indicates one standard deviation from the mean.}
    \label{fig:v-entropy}
\end{figure}
\subsection{Information Flow Analysis}
We analyze the flow of contextual information using models of similar sizes (Figure~\ref{fig:entropy_cf}), as well as the Llama series models of varying sizes (Figure~\ref{fig:entropy_nq}). The details of the experimental settings are provided in Appendix~\ref{appendix:sec2_exp}.

As shown in the Figure~\ref{fig:v-entropy}, the Llama models generally exhibit relatively higher values of $\mathcal{V}$-usable information in their intermediate layers than the final layer. A comparison between Figure~\ref{fig:entropy_cf} and \ref{fig:entropy_nq} reveals that the models exhibit a characteristic information fluctuation across the intermediate layers. Specifically, a subset of layers maintains a high, monotonically increasing \( \mathcal{V} \)-usable information, followed by either a plateau or a decrease in the deeper layers.

The analysis reveal that \textbf{the $\mathcal{V}$-usable information does not follow a monotonically increasing trend toward the output layer}. 
Therefore, we propose leveraging the contextual information within the internal states of LLMs to maintain a continuous growth trend, which may potentially counteract subsequent degradation (or stagnation) effects. 

\section{CaLE: Context-Aware Layer Enhancement}
During the inference process, information fluctuations can occur with degradation (or stagnation). This leads to a reduction in the amount of contextual information at the final layer~\cite{skean}, resulting in a loss of contextual faithfulness during the final decoding.

To mitigate this issue, we propose CaLE, which first identifies the context-aware layers before degradation and then enhances the contextual representations within these layers. This improvement helps to elevate the $I_{\mathcal{V}}(\boldsymbol{h}_f; Y)$ in the final layer, thereby enhancing the model's faithfulnes. Furthermore, we provide theoretical proofs to guarantee the effectiveness of CaLE.

\subsection{Layer Enhancement Methods} \label{sec:theory}
According to Formula~\ref{eq:def_v}, \( I_{\mathcal{V}}(\boldsymbol{h}_f; Y) \) can be maximized by minimizing the \( H_\mathcal{V}(Y|\boldsymbol{h}_f) \) of the final layer. To minimize the $\mathcal{V}$-entropy, we propose two intervene methods for enhancing the layer with rich contextual knowledge:

\paragraph{\underline{A}mplification of Representations at Layer $l$ (CaLE-A).}
For layer \( l \), the representation \( \boldsymbol{h}_{l} \) is directly amplified by a factor \( \alpha_1 \). The enhanced representation \( \boldsymbol{h}^{'}_{l} \) is given by:
\begin{equation}
    \boldsymbol{h}^{'}_{l} = \alpha_1 \cdot \boldsymbol{h}_l
\end{equation}
where \( \alpha_1 \) is a hyperparameter that amplifies the representation of layer \( l \).

\paragraph{\underline{R}esidual Connection from Layer \( l \) to Subsequent Layers (CaLE-R).}
A residual connection is added from layer \( l \) to the representations of $\alpha_2$ subsequent layers. For any layer \( k \) where \( l+1 \leq k \leq l+\alpha_2 \), the enhanced representation \( \boldsymbol{h}^{'}_{l} \) is computed as:
\begin{equation}
    \boldsymbol{h}^{'}_{k} = \boldsymbol{h}_{k} + \boldsymbol{h}_{l}
\end{equation}
where \( \boldsymbol{h}_{l} \) is the representation of layer \( l \), and \( \boldsymbol{h}_{k} \) is the original representation of layer \( k \). 

Both methods ensure that contextual information at layer $l$ is enhanced and accumulated across subsequent layers.

\subsubsection{Theoretical Support for Enhancement}
In this section, we present theoretical support for the enhancement on a context-aware layer to minimize the conditional entropy $H_\mathcal V(Y|\boldsymbol{h}_f)$.

First, we expand the function $f[\cdot]$ at the final layer through the lens of residual stream~\cite{elhage2021mathematical, induction-head}:
\begin{equation}
    f[\boldsymbol{h}_{f}](Y) = \frac{e^{ v_k + u_k(v_k)}}{\sum_{j\in {|Vocab|}} e^{ v_j + u_j(v_j)}}
\end{equation}
where $v$ denotes the logits at layer $l$, and $ u(v) $ includes logit contributions from layer $l+1$ to the final layer $f$. The deduction is detailed in Appendix~\ref{appendix:residual}.

At final layer, the minimization objective $H_\mathcal V(Y|\boldsymbol{h}_f)$ (defined as $H_{\mathcal V_f}$) can be simplified into the following form:
\begin{equation}
    H_{\mathcal V_f}
    =\mathbb{E}[\ln \sum_j e^{v_j + u_j(v_j)} - (v_k + u_k(v_k))]
\end{equation}

As derived in detail in Appendix~\ref{appendix:amplification}, the $H_{\mathcal V_f}$ with layer $l$ amplification is given by:
\begin{equation} \label{eq:h_alpha}
    H_{\mathcal V_f}(\alpha) \defeq \mathbb{E}[\ln \sum_j e^{\alpha v_j + u_j(\alpha v_j)} - \left(\alpha v_k + u_k(\alpha v_k)\right)]
\end{equation}
Additionally, Appendix~\ref{appendix:residual connextion} demonstrates that the residual method is equivalent to a specific value of $\alpha$\footnote{Refer to Formula~\ref{eq:p_modified1} and~\ref{eq:p_modified2} in Appendix~\ref{appendix:enhancement}}, thereby the theoretical framework is also applicable to the residual method.

\begin{proposition}\label{proposition}\textnormal{[Proof in Appendix~\ref{appendix:proof}]}
Let $\alpha$ denote the amplification factor applied to the hidden states at this layer. 
If $k=\arg\max_j v_j$, then 
\begin{equation}
    \lim_{\alpha \to \infty} H_{\mathcal V_f}(\alpha) \approx 0
\end{equation}
\end{proposition}

Since we cannot guarantee that \( v_k \) will achieve the maximum probability proportion at a specific layer, we propose setting \( \alpha > 1 \) as a fixed hyperparameter for the enhancement method. This adjustment amplifies the probabilities of top-ranking tokens while proportionally attenuating the noise from less relevant tokens. In our experiments, we discuss the impact of different values of \( \alpha \).

\subsection{Identifying the Context-Aware Layer}\label{sec:method}
CaLE amplifies the flow of contextual information at an appropriate layer, which can produce significant performance benefits.
In this section, we describe the identification method for the layer.

\subsubsection{Supervised Layer Identification}
The supervised method involves selecting an optimal lay within the Transformer model, by evaluating model performance on a validation set.

Given a set of candidate layers \(\mathcal{L} = \{l_1, l_2, \dots, l_{n-1}\}\), the method computes the performance \(A(l_i, D_{\text{val}})\) for each layer \(l_i\) on the validation set \(D_{\text{val}}\). The optimal layer \(l^*\) is identified as the one that maximizes validation accuracy, formally expressed as:
\begin{equation}
    l^* = \arg\max_{l_i \in \mathcal{L}} A(l_i, D_{\text{val}})
\end{equation}
Subsequently, the selected layer \(l^*\) is used to evaluate the model's performance on the test set \(D_{\text{test}}\), yielding the final test performance \(A_{\text{test}} = A(l^*, D_{\text{test}})\).

\subsubsection{Unsupervised Layer Identification}
In real-world scenarios, label $Y$ may not be available for evaluating layer enhancement performance. We aim to approximate $\mathcal{V}$-usable information through an alternative metric.

Since the answer $Y$ is uniquely determined by the context-query pair $(c,q)$, the information content encoded in $(c,q)$ necessarily exceeds that of $Y$. Then we can establish the following:
\begin{equation} \label{eq:dpi}
    I_\mathcal V(\boldsymbol{h}_l; Y) \leq I_\mathcal V(\boldsymbol{h}_l; c, q)
\end{equation}
where
\begin{equation}
    I_\mathcal V(\boldsymbol{h}_l; c,q) = I_\mathcal V(\boldsymbol{h}_l;q) + I_\mathcal V(\boldsymbol{h}_l; c \mid q)
\end{equation}
Based on the relationship between Kullback-Leibler (KL) divergence~\cite{kl} and MI, we have:
\begin{align} \label{eq:kl}
    &\quad I_\mathcal V(\boldsymbol{h}_l; Y) \leq I_\mathcal V(\boldsymbol{h}_l; q) + I_\mathcal V(\boldsymbol{h}_l; c\mid q)  \nonumber \\
    &=\mathbb{E}_{P(q)} \Big[ \mathrm{KL}[ P(\boldsymbol{h}_l \mid q) \,\|\, P(\boldsymbol{h}_l) ]\Big] + \nonumber\\
    & \quad\  \mathbb{E}_{P(q, c)} \Big[ \mathrm{KL}[ P(\boldsymbol{h}_l \mid q, c) \,\|\, P(\boldsymbol{h}_l \mid q) ]\Big] 
\end{align}
It suggests that we can estimate the upper bound of $\mathcal V$-usable information through the KL divergences of the distribution of hidden states.

\paragraph{Layer Identification based on KL Divergence.} \label{sec:kl}
Given that this approximation imposes only a unidirectional constraint, it does not provide a definitive guarantee for the \( \mathcal{V} \)-usable information. Therefore, we conduct empirical statistics to assess the reliability of the KL divergence as a measurement criterion.

We denote the KL divergences in Formula~\ref{eq:kl} as follows:
\begin{align}
    KL_q(l) &\defeq \text{KL}[ P(\boldsymbol{h}_l \mid q) \,\|\, P(\boldsymbol{h}_l) ] \label{eq:kl_c}\\
    KL_c(l) &\defeq \text{KL}[ P(\boldsymbol{h}_l \mid q, c) \,\|\, P(\boldsymbol{h}_l \mid q) ]
\end{align}
We estimate the KL divergences on the CounterFact~\cite{MengBAB22} dataset with different models. The settings are detailed in Appendix~\ref{appendix:se3_exp}.

\begin{figure}[t] 
    \centering
    \begin{subfigure}{0.23\textwidth}
        \includegraphics[width=\textwidth]{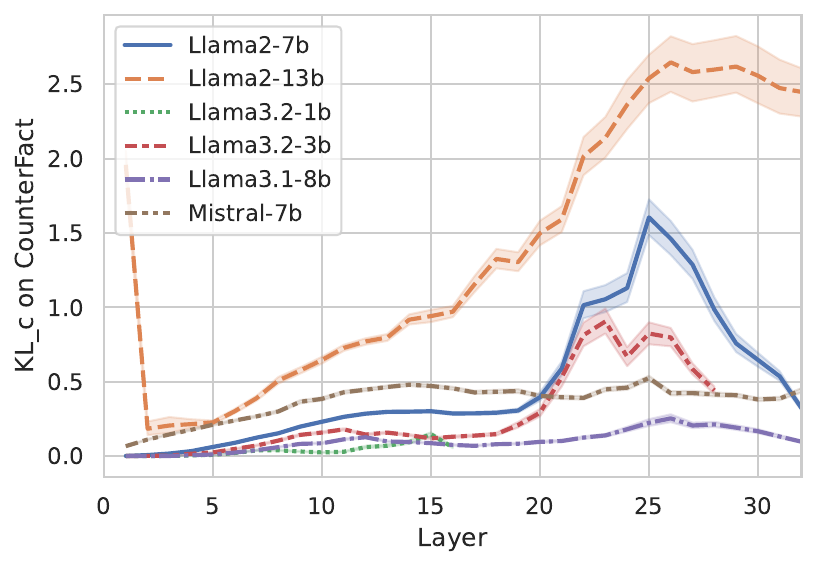}
        \caption{\( \mathrm{KL}_{l_c} \) variation}
        \label{fig:kl_c}
    \end{subfigure}
\hfill
    \begin{subfigure}{0.23\textwidth}
        \includegraphics[width=\textwidth]{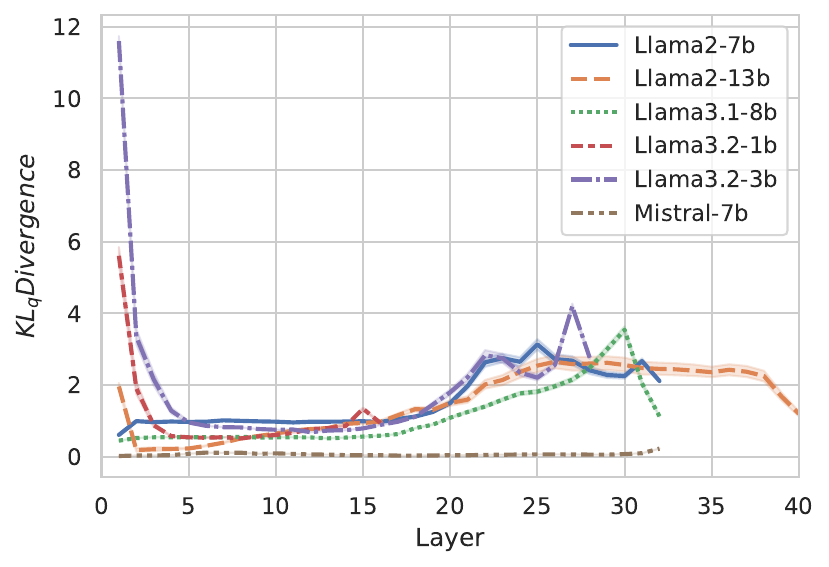}
        \caption{\( \mathrm{KL}_{l_q} \) variation}
        \label{fig:kl_q}
    \end{subfigure}
    \caption{Variation of the KL divergences across layers in different models. The $\mathrm{KL}_q$ quantifies the impact of question conditioning on layer representations by measuring their distributional divergence, while $\mathrm{KL}_c$ captures the incremental influence of context conditioning given the question on these representations. The shaded region represents the confidence interval.}
    \label{fig:kl}
\end{figure}
Figure~\ref{fig:kl_c} demonstrates a strong correlation between \( KL_c(l) \) and \( I_{\mathcal{V}}(\boldsymbol{h}_l; Y) \) across the models. This consistency suggests that the divergence can effectively approximate the \( \mathcal{V} \)-usable information, capturing the incremental influence of context conditioning given the question. Furthermore, for the same question, the information of the correct answer is inherently contained within the context, making \( KL_q \) relatively irrelevant. Supporting evidence from Figure~\ref{fig:kl_q} indicates a low correlation between \( I_{\mathcal{V}}(\boldsymbol{h}_l; Y) \) and \( KL_q \). Therefore, we propose the following approximation:
\begin{equation} \label{eq:v-kl}
    I_{\mathcal{V}}(\boldsymbol{h}_l; Y) \propto\mathbb{E}_{P(q, c)} \text{KL} \left[ P(\boldsymbol{h}_l \mid q, c) \,\|\, P(\boldsymbol{h}_l \mid q) \right]
\end{equation}

We identify the optimal layer by selecting one that exhibits maximal information in \( I_{\mathcal{V}}(\boldsymbol{h}_l; Y) \), which is measured by \( KL_c(l) \) according to Formula~\ref{eq:v-kl}. Therefore, the layer is selected as follows:
\begin{align}
l^* = \arg\max_{l} \; \mathbb{E}_{P(q,c)} [ \mathrm{KL}_c(l)] 
\end{align}
Due to the term $\mathbb{E}_{P(q, c)}$, we measure the average $\mathrm{KL}_c$ across all data points for layer selection. 
\section{Experiments}\label{exp:experiments}
\addtolength{\tabcolsep}{-0.3em}
\begin{table*}[t]
    \tiny
    \centering
    \resizebox{0.95\textwidth}{!}{
    \begin{threeparttable}
    \begin{tabular}{ll|ccccc|ccccc|ccccc}
         \cmidrule[1pt]{1-17}
         && \multicolumn{5}{c}{\textbf{Total}} &\multicolumn{5}{c}{\textbf{Unknown}}  &\multicolumn{5}{c}{\textbf{Known}} \\
         \cmidrule(lr){3-7}\cmidrule(lr){8-12}\cmidrule(lr){13-17}
          \textbf{Models}& \textbf{Methods}& Original & Early Exit & IRCAN & CaLE-R & CaLE-A & Original & Early Exit & IRCAN & CaLE-R & CaLE-A & Original & Early Exit & IRCAN & CaLE-R & CaLE-A \\
         \midrule
         \multirow{4}{*}{\textbf{Llama2-7B}}& Regular* & 54.32 & 70.76 & 71.15 & \cellcolor{blue!5}74.86 & \cellcolor{blue!5}\textbf{74.98} & 42.53 & 66.47 & 63.56 & \cellcolor{blue!5}69.22 & \cellcolor{blue!5}\textbf{69.62}  & 90.90  & 86.86 & 89.35 & \cellcolor{blue!5}\textbf{92.25}  & \cellcolor{blue!5}91.30\\
         &{CD}     &{59.22} & {71.36} &{74.41} &\cellcolor{blue!5}\textbf{76.56} & \cellcolor{blue!5}{76.26} & {51.12} & {67.07} & {69.57} & \cellcolor{blue!5}71.41 & \cellcolor{blue!5}{\textbf{71.66}} & {83.51} & {86.21} & {90.35} & \cellcolor{blue!5}\textbf{92.15} & \cellcolor{blue!5}{90.75}  \\
         &{CAD}    &{63.67} &{69.92} &{69.35} &\cellcolor{blue!5}\textbf{76.81} &\cellcolor{blue!5}{75.66} & {57.92} & {66.32} & {66.17} & \cellcolor{blue!5}\textbf{72.21} & \cellcolor{blue!5}71.44 & {79.51}  & {84.23}  & {77.66} & \cellcolor{blue!5}\textbf{89.41}& \cellcolor{blue!5}{88.66}   \\
         &{COIECD} &{63.72} &{70.26} &{69.62} &\cellcolor{blue!5}\textbf{76.76} &\cellcolor{blue!5}{75.81} & {57.47}& {66.62} & {66.87}  & \cellcolor{blue!5}\textbf{71.91} & \cellcolor{blue!5}{71.51} & {81.06} & {84.96} & {77.91} & \cellcolor{blue!5}\textbf{90.35}& \cellcolor{blue!5}{89.26} \\
         \cmidrule(lr){1-17}
         \multirow{4}{*}{\textbf{Llama3.1-8B}}& Regular &57.72 &67.52 &65.57 &\cellcolor{blue!5}71.06 &\cellcolor{blue!5}\textbf{73.91}  & 45.98 & 59.97 & 56.32 & \cellcolor{blue!5}63.57 & \cellcolor{blue!5}\textbf{67.22} & 90.05 & 86.41 & 90.50 &\cellcolor{blue!5}\textbf{92.50} & \cellcolor{blue!5}92.00 \\
         &{CD}     &{62.42} &{67.76} & {67.77} & \cellcolor{blue!5}73.66 &\cellcolor{blue!5}{\textbf{75.46}} & {54.07} & {61.22} & {61.37} & \cellcolor{blue!5}67.42 & \cellcolor{blue!5}{\textbf{70.21}} & {83.96} & {84.56} & {86.46} & \cellcolor{blue!5}89.86 & \cellcolor{blue!5}{\textbf{92.03}}\\
         &{CAD}    &{66.92} &{68.37} &{70.06} &\cellcolor{blue!5}77.16 &\cellcolor{blue!5}{\textbf{77.71}}& {60.87} & {62.57} & {65.22} & \cellcolor{blue!5}{72.66} & \cellcolor{blue!5}\textbf{73.36} & {81.61} & {81.18}  & {82.66}  & \cellcolor{blue!5}\textbf{86.31}& \cellcolor{blue!5}{86.21}  \\
         &{COIECD} &{66.62} &{68.45} &{70.46} &\cellcolor{blue!5}76.91 &\cellcolor{blue!5}{\textbf{77.31}} & {60.02} & {62.42} & {65.02} & \cellcolor{blue!5}71.96 & \cellcolor{blue!5}{\textbf{72.71}} & {82.81} & {82.06} & {84.11} & \cellcolor{blue!5}\textbf{87.16}& \cellcolor{blue!5}{87.06} \\
         \cmidrule(lr){1-17}
         \multirow{4}{*}{\textbf{Llama3.2-3B}}& Regular &57.67 &71.91 &76.61 &\cellcolor{blue!5}76.81 &\cellcolor{blue!5}\textbf{79.21} & 48.28 & 66.07  & 71.71 & \cellcolor{blue!5}71.16 & \cellcolor{blue!5}\textbf{74.31} & 91.95& 91.25 & 93.75 & \cellcolor{blue!5}95.80 & \cellcolor{blue!5}\textbf{95.90}  \\
         &{CD}&{63.02} &{72.71} &{78.46}&\cellcolor{blue!5}79.16 &\cellcolor{blue!5}{\textbf{80.31}}  & {56.77} & {67.37} & {74.56} & \cellcolor{blue!5}74.26  & \cellcolor{blue!5}{\textbf{75.86}} & {85.96} & {89.36}  & {93.50}& \cellcolor{blue!5}\textbf{94.65}& \cellcolor{blue!5}{94.35}  \\
         &{CAD} &{69.77} &{73.52}&{77.71} &\cellcolor{blue!5}81.11 &\cellcolor{blue!5}{\textbf{82.06}}  & {64.87} & {68.85} & {74.16} & \cellcolor{blue!5}77.01 & \cellcolor{blue!5}{\textbf{78.26}} & {84.36} & {85.81} & {90.02} & \cellcolor{blue!5}91.50 & \cellcolor{blue!5}{\textbf{92.05}} \\
         &{COIECD} &{69.12} &{73.69} &{78.52}&\cellcolor{blue!5}80.81 &\cellcolor{blue!5}{\textbf{81.81}}  & {64.02} & {68.73} & {74.65} & \cellcolor{blue!5}76.66 & \cellcolor{blue!5}{\textbf{78.01}} & {85.51} & {86.96} & {91.30}& \cellcolor{blue!5}92.15 & \cellcolor{blue!5}{\textbf{92.60}} \\

        \bottomrule[1pt]
    \end{tabular}
     \begin{tablenotes}
        \item[*] Regular refers to the greedy decoding strategy.
    \end{tablenotes}
     \end{threeparttable}
    }
    \vspace{-3mm}
    \caption{EM results on the CounterFact dataset with \textbf{supervised} intervene methods. The CaLE-A/R method denote the amplification or residual methods for enhancement. The highest scores with different decoding strategies are highlighted in \textbf{bold}.}
    \label{tab:main-results}
    \vspace{-3mm}
\end{table*}
\addtolength{\tabcolsep}{0.3em}

\subsection{Settings}
\paragraph{Data.} 
We evaluate the performance of CaLE across diverse QA datasets, including CounterFact~\cite{MengBAB22} NQ~\cite{NQ_KwiatkowskiPRCP19}, SQuAD~\cite{squad}, and StrategyQA~\cite{strategyqa}. \footnote{For the CounterFact, $\alpha_1=5$ for CaLE-A, and $\alpha_2=3$ for CaLE-R. For the other datasets, $\alpha_1=3$ and $\alpha_2=1$.} More details of the data are in Appendix~\ref{appendix:data_detail}.

\paragraph{Models.} We conduct experiments on state-of-the-art language models, including several variants from the Llama model family—specifically, Llama2-7B, Llama3.1-8B, and Llama3.2-3B—as well as the Mistral-7B and Gemma2-9B models.

\paragraph{Baselines.} 
To demonstrate the effectiveness of CaLE, we compare it with the following baselines: \textbf{Original}, which refers to the LLMs without any modification; \textbf{Early Exit}~\cite{DBLP:conf/eacl/XinTYL21,bi_values, fansq}, where the model exits early at the layer with the best performance; and \textbf{IRCAN}~\cite{IRCAN}, which reweights the neurons critical for processing contextual cues. Both intervention methods are supervised.

For the supervised CaLE method, we construct the validation set using 0.5k samples, randomly selected from the training data to ensure no overlap with the test set. 

We also combine several decoding methods, since the above methods work in completely different ways with decoding strategies: Context-Aware Decoding (\textbf{CAD})~\cite{ShiHLTZY24}, Contrastive Decoding (\textbf{CD})~\cite{LiHFLEHZL23}, and Contextual Information-Entropy Constraint Decoding (\textbf{COIECD})~\cite{YuanYWLZL24}. 
Detailed description are provided in Appendix~\ref{appendix:decoding}.

\paragraph{Metrics.} We use the Exact Match (EM) and F1 scores for evaluating the QA performance of LLMs. For the binary classification in StrategyQA, the accuracy is used as the metric.

\subsection{A Thorough Analysis on the CounterFact Dataset}
We first conduct a comprehensive analysis by applying supervised CaLE intervention on the CounterFact dataset. Specifically, we partition the CounterFact dataset into "known" and "unknown" subsets (with "total" representing the complete set). The classification is based on whether the external contextual knowledge is consistent with the model's parametric knowledge~\cite{RenWQZ00W025}\footnote{The details of the posteriori judgement for the dataset are in Appendix~\ref{appendix:prior}.}.

\subsubsection{Overall Performance}
\paragraph{Superior Performance.}
As shown in Table~\ref{tab:main-results}, the experimental results demonstrate that both supervised CaLE variants (CaLE-A and CaLE-R) consistently outperform baseline methods across all models. This suggests that enhancing the context-aware layer within the model significantly improves context-faithfulness generation. Furthermore, the advantage of our method is particularly pronounced when handling new ("unknown") knowledge, whereas ICRAN underperforms even compared to Early Exit method on Llama2-7b and Llama3.1-8b with Regular decoding strategy.
\begin{figure}[t] 
    \centering
    \begin{subfigure}{0.22\textwidth}
        \includegraphics[width=\textwidth]{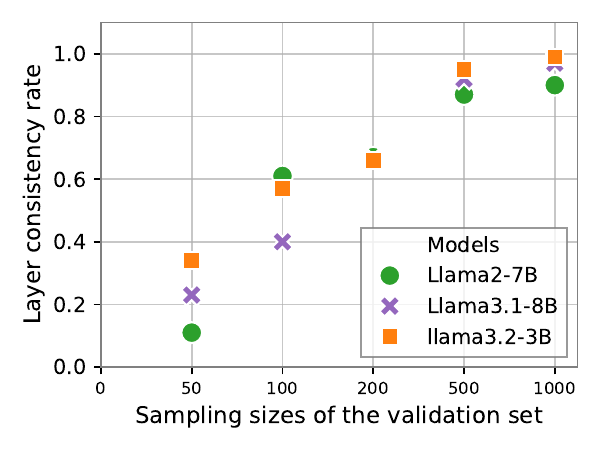}
        \caption{Layer consistency rate across 20 sampling trials for different sampling sizes of the validation set.}
        \label{fig:dev_sample}
    \end{subfigure}
\hfill
    \begin{subfigure}{0.24\textwidth}
        \includegraphics[width=\textwidth]{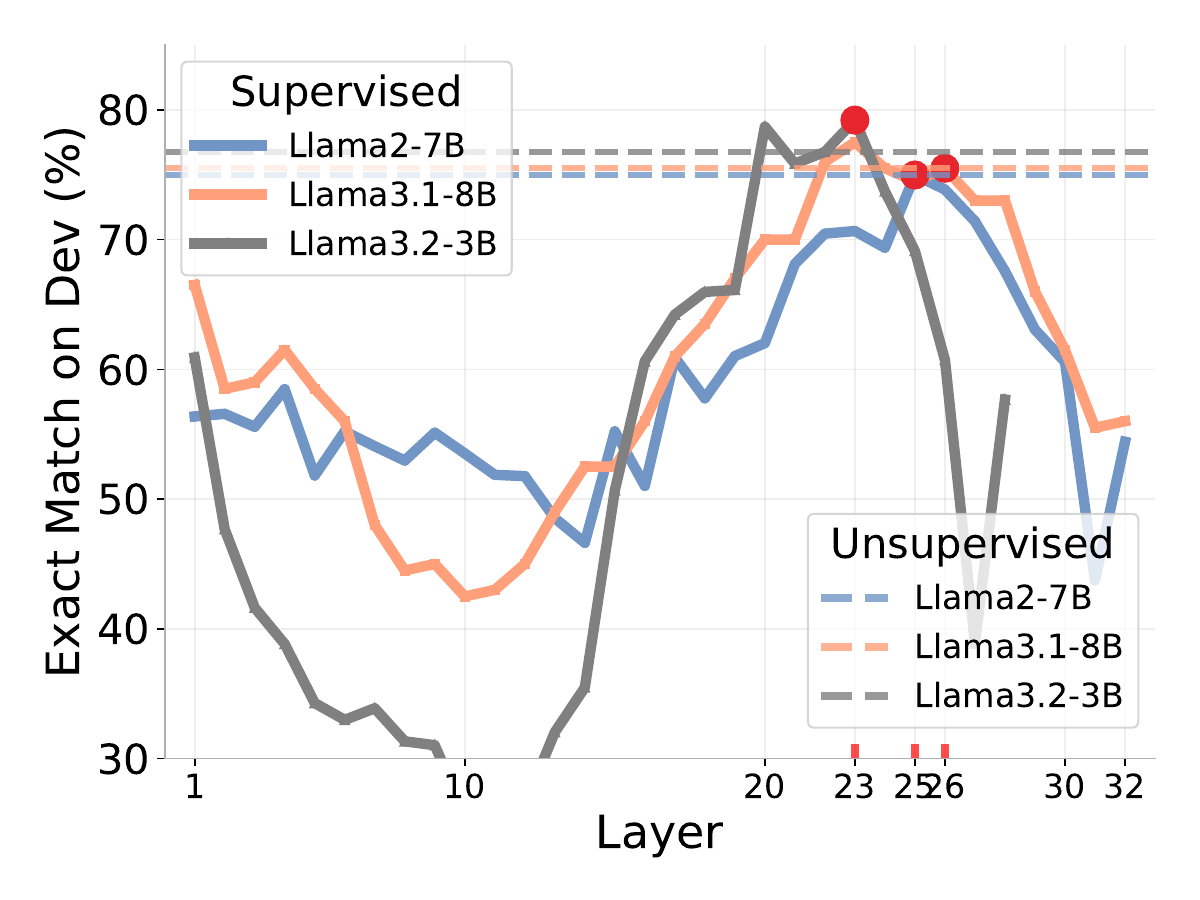}
        \caption{Layer selection comparison between supervised and unsupervised CaLE-A.}
        \label{fig:unsupervised}
    \end{subfigure}
    \caption{Validation set size impact on supervised layer selection and comparative layer selection with unsupervised CaLE. The selected layers are detailed in Table~\ref{tab:layer}.}
\end{figure}
\begin{table}[t]
\centering
\resizebox{0.48\textwidth}{!}{
\begin{tabular}{l|cc|cc|cc}
\cmidrule[1pt]{1-7}
EM & \multicolumn{2}{c}{Llama2-7B} & \multicolumn{2}{c}{Llama3.1-8B} & \multicolumn{2}{c}{Llama3.2-3B} \\
\cmidrule(lr){2-3}\cmidrule(lr){4-5}\cmidrule(lr){6-7}
$\Delta$=(\textbf{sup.- unsup.}) & CaLE-R($\Delta$) & CaLE-A($\Delta$)  & CaLE-R($\Delta$) & CaLE-A($\Delta$)  & CaLE-R($\Delta$) & CaLE-A($\Delta$) \\
\midrule
\textbf{Total}  & 74.86$\db{-0.00}$ & \textbf{74.98}$\db{-0.00}$  & 71.06$\db{-0.56}$ & \textbf{73.91}$\db{-1.23}$ & 76.81$\db{-0.00}$ & \textbf{79.21}$\db{-0.00}$ \\
\textbf{Unknown}  & 69.22$\db{-0.00}$ & \textbf{69.62}$\db{-0.00}$ & 63.57$\db{-1.05}$ & \textbf{67.22}$\db{-1.14}$ & 71.16$\db{-0.00}$ &\textbf{74.31}$\db{-0.00}$ \\
\textbf{Known} &  \textbf{92.25}$\db{-0.00}$ & 91.30$\db{-0.00}$ & \textbf{92.50}$\db{-0.50}$ & 92.00$\db{-0.91}$ & 95.80$\db{-0.00}$ & \textbf{95.90}$\db{-0.00}$ \\
\bottomrule
\end{tabular}
}
\caption{Performance comparison between supervised and unsupervised CaLE. The black numbers represent the scores of the supervised CaLE method, with the values in "()" indicating the difference between the supervised and unsupervised methods.}
\label{tab:comparison}
\end{table}
\paragraph{Difference Between CaLE-R and CaLE-A}
While both CaLE-R and CaLE-A enhance accuracy, their mechanisms lead to differences in performance. CaLE-R, which incorporates residual connections, provides a stable but modest improvement in the "Unknown" subset. In contrast, CaLE-A, which amplifies knowledge representations, achieves nearly the highest scores across all models. This indicates that CaLE-A’s amplification mechanism is more effective at handling new factual knowledge. On the other hand, CaLE-R excels in the generation of consistent internal and external knowledge, as evidenced by its performance in the "Known" subset.

\paragraph{Versatility with Decoding Methods.}
One of the key strengths of CaLE lies in its versatility across different decoding methods. Regardless of the strategy used—CD, CAD, or COIECD—CaLE-based models consistently achieve higher EM scores compared to other baselines. In contrast, Early Exit and IRCAN do not show the same level of reliability, with fluctuating gains and occasional declines, particularly in the Llama2-7B model.

\subsubsection{Comparison between unsupervised and supervised methods of CaLE}
In the case of supervised CaLE, we use a validation set size of 0.5k. As shown in Figure~\ref{fig:dev_sample}, the vertical axis represents the proportion of the mode of the best layer selected based on validation accuracy across 20 trials. The figure indicates that a validation set size of 0.5k is sufficient for robust layer selection.
Next, we analyze the KL-based unsupervised CaLE method on CounterFact dataset. 

\paragraph{Layer Selection Comparison.}
As illustrated in Figure~\ref{fig:unsupervised}, the solid line represents the performance of different amplified layer in a validation set trial. The red dots indicate the layers selected using the KL-based metric in an unsupervised manner. Notably, these layers correspond closely to the peaks in supervised performance, particularly around layers 23, 25, and 26. This strong correlation suggests that the KL-based selection method effectively identifies context-aware layers that contribute significantly to contextual information.

\paragraph{Performance Comparison.} The experimental results in Table~\ref{tab:comparison} further validate this observation. Across various Llama models, although the unsupervised method generally performs worse than the supervised method, it consistently outperforms all other baselines in Table~\ref{tab:main-results}. Notably, for Llama2-7B and Llama3.2-3B, the unsupervised CaLE method achieves scores identical to its supervised counterpart, as both methods identify the same layer for enhancement. These findings underscore the effectiveness of the KL-based CaLE, demonstrating its ability to approximate optimal layer selections without the need for labeled supervision.

\begin{figure}[t] 
    \centering
    \begin{subfigure}{0.23\textwidth}
        \includegraphics[width=\textwidth]{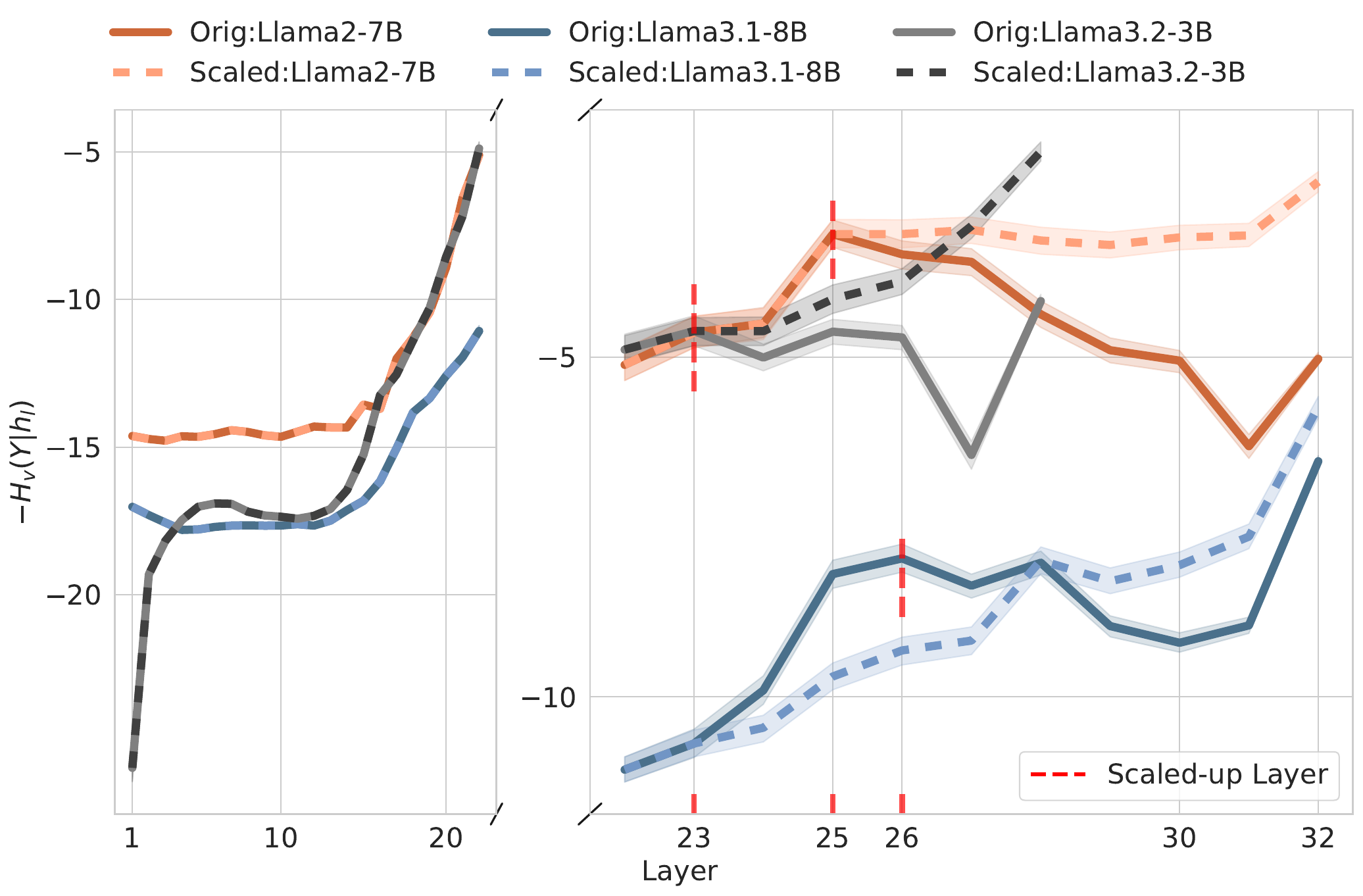}
        \caption{CaLE increases the $\mathcal{V}$-usable information.}
        \label{fig:delta_h}
    \end{subfigure}
\hfill
    \begin{subfigure}{0.23\textwidth}
        \includegraphics[width=\textwidth]{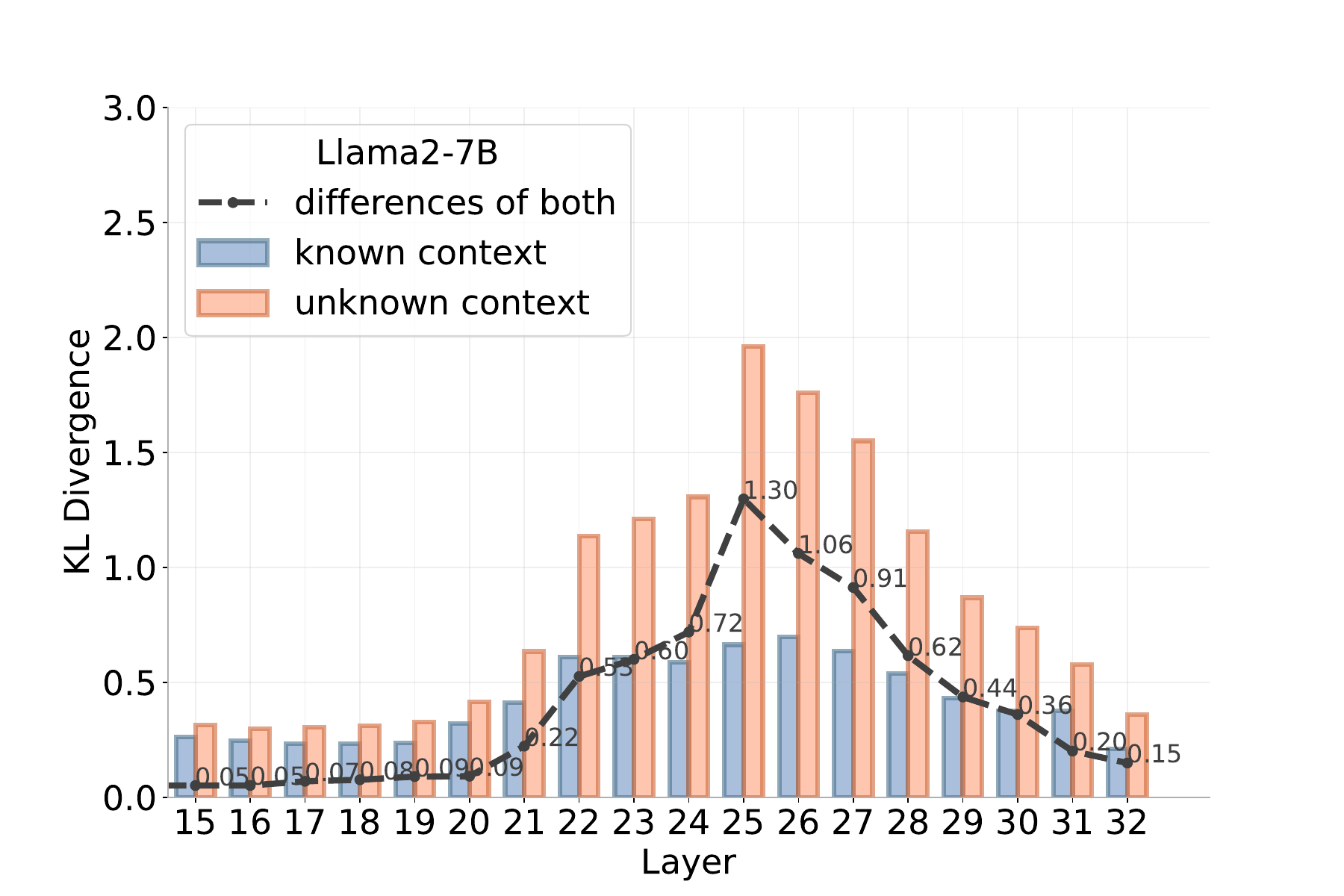}
        \caption{KL Divergence in unknown and known contexts.}
        \label{fig:7b_kl}
    \end{subfigure}
    \caption{Visualization of Analysis on the CounterFact Dataset for Llama  models.}
\end{figure}

\subsubsection{Further Analysis}
\paragraph{Increased Information.}
Figure~\ref{fig:delta_h} provides theoretical validation for the effectiveness of our CaLE-A method. The approach aims to enhance contextual information representation in the model's final layer, quantitatively assessed through negative \( \mathcal{V} \)-entropy measurements. The results demonstrate that the amplification mechanism successfully transforms the previously observed information degradation, represented by solid lines, into an upward trend, as shown by the dashed lines.

\paragraph{Unknown Contexts with Higher KL Divergence.} 
In Figure~\ref{fig:7b_kl}, the KL divergence of contexts in "Unknown" subset consistently exhibits a greater magnitude compared to known contexts across deeper layers. The peak observed at layer 25 aligns with the layer selected by the CaLE method, offering robust validation for both approaches. The KL metric provides an interpretable rationale for layer selection decisions, as it quantifies the distributional impact of the contextual input \( c \) on each layer. This metric effectively captures the extent to which different layers encode and propagate contextual knowledge, particularly for new knowledge.

\subsection{Application on Diverse QA Datasets}
\paragraph{Conflicting Contexts.}To further explore CaLE's effectiveness in handling novel information, we conduct comprehensive contrastive analyses using the NQ dataset~\cite{NQ_KwiatkowskiPRCP19} and its variant, NQ-Swap~\cite{LongprePCRD021}. The NQ-Swap dataset, derived from the original NQ, exclusively consists of conflicting contextual knowledge that contradicts the model's parametric knowledge.

As illustrated in Table~\ref{tab:nq}, the improvement of CaLE is particularly evident when evaluated on the NQ-Swap dataset, which is entirely composed of conflicting knowledge. These findings indicate that CaLE intervention effectively facilitates the utilization of new knowledge in the model.

\begin{table} [t]
\tiny
\centering
\resizebox{\columnwidth}{!}{%
\begin{threeparttable}
\begin{tabular}{l|cc|cc|cc|c}
 \toprule
   & \multicolumn{2}{c}{\textbf{NQ}} & \multicolumn{2}{c}{\textbf{NQ-Swap}}& \multicolumn{2}{c}{\textbf{SQuAD}} & \multicolumn{1}{c}{\textbf{StrategyQA}}  \\
  \cmidrule(lr){2-3} \cmidrule(lr){4-5} \cmidrule(lr){6-7} \cmidrule(lr){8-8}  
    &  EM & F1 &  EM & F1 &  EM & F1 & Acc\\
\midrule
Llama2-7B & 75.84 & 77.48  & 53.73  & 54.92 & 61.37 & 73.02 & 80.41\\
\textcolor{orange}{$\mathbf{unsup.}$}   &——&——&——&——&——&——&———— \\
\phantom{0}+ CaLE-R          & 77.69 & 79.41 & 58.68 & 59.98 & 63.52 & 74.68 & 82.74 \\
\phantom{0}+ CaLE-A          & \textbf{78.19} & \textbf{80.06} & 58.78 & 60.01 & 63.59 & 74.70 & 82.91\\
\textcolor{cyan}{$\mathbf{sup.}$}    &——&——&——&——&——&——&————     \\
\phantom{0}+ IRCAN          & 75.79 & 78.81 & 58.32 & 60.34 & 62.01 & 73.97 & 80.41\\
\phantom{0}+ CaLE-R          & 77.69 & 79.41 & 59.78 & 61.35 & 64.42 & 75.27 & 81.26 \\
\phantom{0}+ CaLE-A          & \textbf{78.19} & \textbf{80.06} & \textbf{63.83} & \textbf{64.83} & \textbf{64.62} & \textbf{75.35} & \textbf{83.16} \\

\midrule
Llama3.1-8B & 76.94 & 78.81  & 49.52  & 50.50 & 64.93 & 78.01 & 85.86 \\
\textcolor{orange}{$\mathbf{unsup.}$}     &——&——&——&——&——&——&————   \\
\phantom{0}+ CaLE-R        & 79.74 & 81.39  & 53.58  & 54.61 & 65.68 & 78.33 & 85.90 \\
\phantom{0}+ CaLE-A          & 79.79 & 81.43 & 53.63 & 54.80 & \textbf{67.38} & \textbf{79.07} & 88.56 \\

\textcolor{cyan}{$\mathbf{sup.}$}    &——&——&——&——&——&——&————    \\
\phantom{0}+ IRCAN          & 79.08 & 80.89 & 59.43 & 60.56 & 64.58 & 76.28 & 87.01\\
\phantom{0}+ CaLE-R          & 78.29 & 80.11 & 56.08 & 57.52 & 65.68 & 78.33 & 86.21 \\
\phantom{0}+ CaLE-A          & \textbf{80.44} & \textbf{81.99} & \textbf{60.52} & \textbf{61.80} & \textbf{67.38} & \textbf{79.07} & \textbf{88.11} \\

\midrule
Mistral-7B & 77.32 & 78.87  & 49.13  & 50.07 & 63.97 & 76.09 & 87.26\\
\textcolor{orange}{$\mathbf{unsup.}$}     &——&——&——&——&——&——&————    \\
\phantom{0}+ CaLE-R        & 79.94 & 80.91  & 53.43  & 54.31 & 65.72 & 77.80 & 87.66 \\
\phantom{0}+ CaLE-A          & \textbf{80.69} & \textbf{81.76} & 53.38 & 54.26 & 65.82 & 77.81 & 88.71 \\

\textcolor{cyan}{$\mathbf{sup.}$}    &——&——&——&——&——&——&————    \\
\phantom{0}+ IRCAN          & 78.61 & 80.03 & 58.02 & 58.82 & 64.27 & 77.02 & 86.76\\
\phantom{0}+ CaLE-R          &  79.94 & 80.91 & 56.38 & 57.30 & 64.12 & 76.36 & 88.01\\
\phantom{0}+ CaLE-A          & \textbf{80.69} & \textbf{81.76} & \textbf{58.58} & \textbf{59.28} & \textbf{66.42} & \textbf{78.01} & \textbf{89.16} \\
\midrule
Gemma2-9B & 78.49 & 81.46 & 47.27 & 49.07 & 61.42 & 75.78 & 84.86\\
\textcolor{orange}{$\mathbf{unsup.}$}     &——&——&——&——&——&——&————    \\
\phantom{0}+ CaLE-R        &  79.54 & 81.93 & 51.76 & 53.84 & 64.22 & 77.06 & 90.01\\
\phantom{0}+ CaLE-A          & 79.75 & 82.12 & 52.33 & 54.25 & 64.92 & 76.93 & 90.47\\
\textcolor{cyan}{$\mathbf{sup.}$}    &——&——&——&——&——&——&————    \\
\phantom{0}+ IRCAN          & 78.74 & 81.39 & \textbf{65.01} & \textbf{66.33} & 64.10 & 76.34 & 84.63\\
\phantom{0}+ CaLE-R          &  79.54 & 81.93 & 62.54 & 64.81 & 66.87 & 78.72 & 90.15\\
\phantom{0}+ CaLE-A          & \textbf{79.94} & \textbf{82.58} & 63.03 & 65.38 & \textbf{67.92} & \textbf{79.88} & \textbf{90.41}\\
 \bottomrule
\end{tabular}
\end{threeparttable}
}
\caption{EM and F1 scores for the diverse QA datasets. The \textcolor{orange}{$\mathbf{unsup.}$} and \textcolor{cyan}{$\mathbf{sup.}$} denote the unsupervised and supervised intervene methods. The best scores are highlighted in \textbf{bold}.}
\label{tab:nq}
\end{table}

\paragraph{Generalization on ComplexQA.}
\begin{figure}[t]
    \centering
    \includegraphics[width=.95\linewidth]{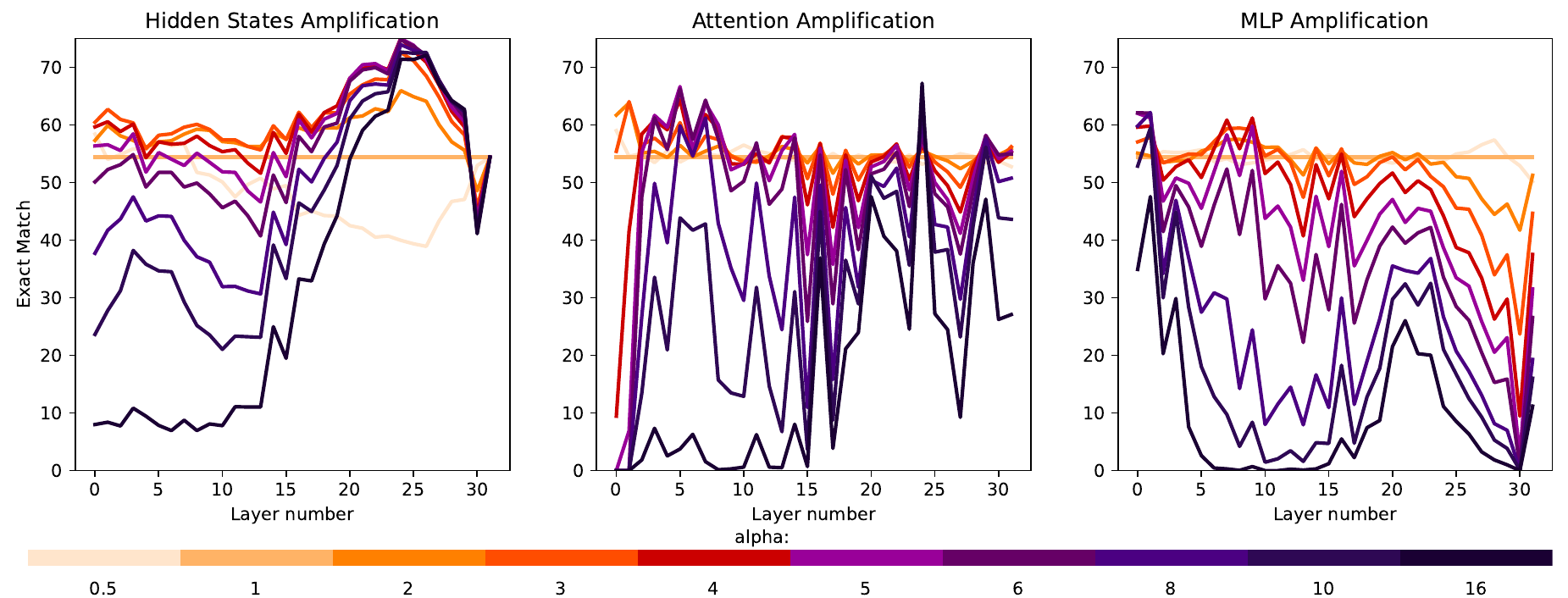}
    \caption{The effect of amplification across different components of Llama2-7b on the CounterFact dataset. The yellow horizontal line is the original EM score evaluated by $\alpha=1$ (without intervention). In the Attention and MLP components, it is clear that the amplification would damage the parameters space as there is very little performance increase. However, for the hidden states, CaLE goes from uniformly harming to improving the performance in the deep layers.}
    \label{fig:ablation}
\end{figure}
We extend our evaluation to other complex QA datasets, including SQuAD~\cite{squad} and StrategyQA~\cite{strategyqa}, across various models. As demonstrated in Table~\ref{tab:nq}, we compare our approach with the strongest baseline method, IRCAN. Our CaLE approach achieves superior performance on the QA datasets compared to the baseline.

For all diverse QA datasets, the CaLE method almostly outperforms the IRCAN method. Specifically, we observe that the IRCAN method does not perform well on the StrategyQA dataset with long contexts, often exhibiting minimal effects. Furthermore, the supervised CaLE method yields better results than its unsupervised counterpart, with CaLE-A outperforming CaLE-R. This suggests that CaLE-A possesses stronger generalization capabilities relative to CaLE-R.

\paragraph{Effectiveness on Different Model Architectures.} Across different model architectures, our method maintains robust performance on the Mistral and Gemma2 models, achieving improvements comparable to those observed with the Llama family models. This finding highlights the effectiveness of our approach.

\subsection{Ablation Study}
In this section, we analyze the effectiveness of amplifying different components of the Llama2-7B model: \textbf{Hidden states}, \textbf{Attention}, and \textbf{MLP} with different values of $\alpha$, as illustrated in Figure~\ref{fig:ablation}.

\paragraph{Value of $\alpha$ .} Amplifying hidden states results in a clear performance boost, with the effectiveness varying across different layers and amplification factors $\alpha$. The optimal amplification factor appears to be between 4 and 6, as evidenced by the higher EM scores in the upper curves of the left plot. 

\paragraph{Intervention Layer.} Notably, the intervention is most effective when applied in later layers (20-25), where all $\alpha$ values lead to convergence around 70\% EM score. This suggests that the model's representational capacity is most malleable and responsive to amplification in these deeper layers, possibly due to their role in high-level feature integration. 

\paragraph{Attention and MLP.} The Attention and MLP amplification show more erratic results, with fluctuating performance across layers. For these components, amplifying either leads to diminishing returns or even decreases in performance, suggesting that these layers do not benefit from amplification in the same way as hidden states.

The poor performance of Attention and MLP amplification can be attributed to several factors. For the attention mechanism, which is finely controlled for inference tasks~\cite{JinCY0XLJ0024,nips/ZhouFZQM24}, further amplification may disrupt the delicate balance, leading to noise amplification. Furthermore, the MLP is generally responsible for storing knowledge~\cite{mlp, mlp2}. The amplifying the entire MLP can result in a proportional increase in all stored knowledge, which may effectively render the amplification meaningless.

\section{Related Work}
Existing approaches to enhancing context-faithfulness in LLMs can be broadly classified into three categories: fine-tuning methods~\cite{bi2024}, external interventions~\cite{conf/emnlp/ZhouZPC23} and internal interventions. The internal interventions predominantly focus on modifying decoding strategies or reweighting knowledge-aware neurons. Methods such as CAD~\cite{ShiHLTZY24} and COIECD~\cite{YuanYWLZL24} optimize decoding strategies through a contrastive mechanism~\cite{LiHFLEHZL23} to promote greater reliance on external information. However, these decoding-based approaches operate at the output level, resulting in only limited improvements. Another stream of research explores the internal states of models. For instance, \citet{JinCY0XLJ0024} and \citet{IRCAN} aim to identify and reweight neurons crucial for processing contextual cues, thereby alleviating conflicts through targeted interventions at critical points.

\section{Conclusion}
In this paper, we propose a novel intervention method called CaLE, which exploits contextual knowledge within LLMs’ internal representations. It strategically amplifies the contextual information growth at an appropriate layer, which facilitates richer representations in the final layer. Our experiments demonstrate that CaLE effectively improves context-faithful generation in QA tasks, particularly in scenarios involving unknown or conflicting contextual knowledge.

\section*{Limitations}

The CaLE approach is to simply conduct one-time intervention during the inference process. While there are several other potential methods to execute interventions continuously, we leave the exploration of these alternatives for future work.

The intervention analysis on the MLP and Attention components adopts the amplification method proposed in the paper. While numerous studies~\cite{mlp,mlp2} have discussed the role of these components in manipulating knowledge, this study does not specifically explore whether alternative intervention methods beyond amplification, or selective interventions within layers, should be employed. Instead, a uniform approach of amplifying entire layers is adopted, which may introduce limitations.
\bibliography{custom}

\begin{thebibliography}{39}
\providecommand{\natexlab}[1]{#1}

\bibitem[{Azaria and Mitchell(2023)}]{abs-2304-13734}
Amos Azaria and Tom~M. Mitchell. 2023.
\newblock The internal state of an {LLM} knows when its lying.
\newblock \emph{CoRR}, abs/2304.13734.

\bibitem[{Bi et~al.(2024)Bi, Huang, Wang, Yang, Zhang, Huang, Mei, Fang, Li, Wei, Deng, Sun, Zhang, and Liu}]{bi2024}
Baolong Bi, Shaohan Huang, Yiwei Wang, Tianchi Yang, Zihan Zhang, Haizhen Huang, Lingrui Mei, Junfeng Fang, Zehao Li, Furu Wei, Weiwei Deng, Feng Sun, Qi~Zhang, and Shenghua Liu. 2024.
\newblock \href {https://arxiv.org/abs/2412.15280} {Context-dpo: Aligning language models for context-faithfulness}.
\newblock \emph{Preprint}, arXiv:2412.15280.

\bibitem[{Chen et~al.(2024)Chen, Liu, Chen, Gu, Wu, Tao, Fu, and Ye}]{chen}
Chao Chen, Kai Liu, Ze~Chen, Yi~Gu, Yue Wu, Mingyuan Tao, Zhihang Fu, and Jieping Ye. 2024.
\newblock {INSIDE:} llms' internal states retain the power of hallucination detection.
\newblock In \emph{The Twelfth International Conference on Learning Representations, {ICLR} 2024, Vienna, Austria, May 7-11, 2024}.

\bibitem[{Elazar et~al.(2021)Elazar, Kassner, Ravfogel, Ravichander, Hovy, Sch{\"{u}}tze, and Goldberg}]{ElazarKRRHSG21}
Yanai Elazar, Nora Kassner, Shauli Ravfogel, Abhilasha Ravichander, Eduard~H. Hovy, Hinrich Sch{\"{u}}tze, and Yoav Goldberg. 2021.
\newblock Measuring and improving consistency in pretrained language models.
\newblock \emph{Trans. Assoc. Comput. Linguistics}, 9:1012--1031.

\bibitem[{Elhage et~al.(2021)Elhage, Nanda, Olsson, Henighan, Joseph, Mann, Askell, Bai, Chen, Conerly et~al.}]{elhage2021mathematical}
Nelson Elhage, Neel Nanda, Catherine Olsson, Tom Henighan, Nicholas Joseph, Ben Mann, Amanda Askell, Yuntao Bai, Anna Chen, Tom Conerly, et~al. 2021.
\newblock A mathematical framework for transformer circuits.
\newblock \emph{Transformer Circuits Thread}, 1(1):12.

\bibitem[{Ethayarajh et~al.(2022)Ethayarajh, Choi, and Swayamdipta}]{EthayarajhCS22}
Kawin Ethayarajh, Yejin Choi, and Swabha Swayamdipta. 2022.
\newblock Understanding dataset difficulty with \emph{V}-usable information.
\newblock In \emph{International Conference on Machine Learning, {ICML} 2022, 17-23 July 2022, Baltimore, Maryland, {USA}}, volume 162 of \emph{Proceedings of Machine Learning Research}, pages 5988--6008. {PMLR}.

\bibitem[{Fan et~al.(2024)Fan, Jiang, Li, Meng, Han, Shang, Sun, Wang, and Wang}]{fansq}
Siqi Fan, Xin Jiang, Xiang Li, Xuying Meng, Peng Han, Shuo Shang, Aixin Sun, Yequan Wang, and Zhongyuan Wang. 2024.
\newblock Not all layers of llms are necessary during inference.
\newblock \emph{CoRR}, abs/2403.02181.

\bibitem[{Gao et~al.(2024)Gao, Xiong, Gao, Jia, Pan, Bi, Dai, Sun, Wang, and Wang}]{gao2024}
Yunfan Gao, Yun Xiong, Xinyu Gao, Kangxiang Jia, Jinliu Pan, Yuxi Bi, Yi~Dai, Jiawei Sun, Meng Wang, and Haofen Wang. 2024.
\newblock \href {https://arxiv.org/abs/2312.10997} {Retrieval-augmented generation for large language models: A survey}.
\newblock \emph{Preprint}, arXiv:2312.10997.

\bibitem[{Geva et~al.(2021{\natexlab{a}})Geva, Khashabi, Segal, Khot, Roth, and Berant}]{strategyqa}
Mor Geva, Daniel Khashabi, Elad Segal, Tushar Khot, Dan Roth, and Jonathan Berant. 2021{\natexlab{a}}.
\newblock Did aristotle use a laptop? {A} question answering benchmark with implicit reasoning strategies.
\newblock \emph{Trans. Assoc. Comput. Linguistics}, 9:346--361.

\bibitem[{Geva et~al.(2021{\natexlab{b}})Geva, Schuster, Berant, and Levy}]{mlp2}
Mor Geva, Roei Schuster, Jonathan Berant, and Omer Levy. 2021{\natexlab{b}}.
\newblock Transformer feed-forward layers are key-value memories.
\newblock In \emph{Proceedings of the 2021 Conference on Empirical Methods in Natural Language Processing, {EMNLP} 2021, Virtual Event / Punta Cana, Dominican Republic, 7-11 November, 2021}, pages 5484--5495. Association for Computational Linguistics.

\bibitem[{Hanna et~al.(2023)Hanna, Liu, and Variengien}]{nips/0001LV23}
Michael Hanna, Ollie Liu, and Alexandre Variengien. 2023.
\newblock How does {GPT-2} compute greater-than?: Interpreting mathematical abilities in a pre-trained language model.
\newblock In \emph{Advances in Neural Information Processing Systems 36: Annual Conference on Neural Information Processing Systems 2023, NeurIPS 2023, New Orleans, LA, USA, December 10 - 16, 2023}.

\bibitem[{Ji et~al.(2023)Ji, Lee, Frieske, Yu, Su, Xu, Ishii, Bang, Madotto, and Fung}]{journals/csur/JiLFYSXIBMF23}
Ziwei Ji, Nayeon Lee, Rita Frieske, Tiezheng Yu, Dan Su, Yan Xu, Etsuko Ishii, Yejin Bang, Andrea Madotto, and Pascale Fung. 2023.
\newblock Survey of hallucination in natural language generation.
\newblock \emph{{ACM} Comput. Surv.}, 55(12):248:1--248:38.

\bibitem[{Jin et~al.(2024)Jin, Cao, Yuan, Chen, Xu, Li, Jiang, Liu, and Zhao}]{JinCY0XLJ0024}
Zhuoran Jin, Pengfei Cao, Hongbang Yuan, Yubo Chen, Jiexin Xu, Huaijun Li, Xiaojian Jiang, Kang Liu, and Jun Zhao. 2024.
\newblock Cutting off the head ends the conflict: {A} mechanism for interpreting and mitigating knowledge conflicts in language models.
\newblock In \emph{Findings of the Association for Computational Linguistics, {ACL} 2024, Bangkok, Thailand and virtual meeting, August 11-16, 2024}, pages 1193--1215. Association for Computational Linguistics.

\bibitem[{Ju et~al.(2024)Ju, Sun, Du, Yuan, Ren, and Liu}]{Ju0DYRL24}
Tianjie Ju, Weiwei Sun, Wei Du, Xinwei Yuan, Zhaochun Ren, and Gongshen Liu. 2024.
\newblock How large language models encode context knowledge? {A} layer-wise probing study.
\newblock In \emph{Proceedings of the 2024 Joint International Conference on Computational Linguistics, Language Resources and Evaluation, {LREC/COLING} 2024, 20-25 May, 2024, Torino, Italy}, pages 8235--8246. {ELRA} and {ICCL}.

\bibitem[{Kullback and Leibler(1951)}]{kl}
S.~Kullback and R.~A. Leibler. 1951.
\newblock On information and sufficiency.
\newblock \emph{The Annals of Mathematical Statistics}, 22(1):79--86.

\bibitem[{Kwiatkowski et~al.(2019)Kwiatkowski, Palomaki, Redfield, Collins, Parikh, Alberti, Epstein, Polosukhin, Devlin, Lee, Toutanova, Jones, Kelcey, Chang, Dai, Uszkoreit, Le, and Petrov}]{NQ_KwiatkowskiPRCP19}
Tom Kwiatkowski, Jennimaria Palomaki, Olivia Redfield, Michael Collins, Ankur~P. Parikh, Chris Alberti, Danielle Epstein, Illia Polosukhin, Jacob Devlin, Kenton Lee, Kristina Toutanova, Llion Jones, Matthew Kelcey, Ming{-}Wei Chang, Andrew~M. Dai, Jakob Uszkoreit, Quoc Le, and Slav Petrov. 2019.
\newblock Natural questions: a benchmark for question answering research.
\newblock \emph{Trans. Assoc. Comput. Linguistics}, 7:452--466.

\bibitem[{Li et~al.(2023)Li, Holtzman, Fried, Liang, Eisner, Hashimoto, Zettlemoyer, and Lewis}]{LiHFLEHZL23}
Xiang~Lisa Li, Ari Holtzman, Daniel Fried, Percy Liang, Jason Eisner, Tatsunori Hashimoto, Luke Zettlemoyer, and Mike Lewis. 2023.
\newblock Contrastive decoding: Open-ended text generation as optimization.
\newblock In \emph{Proceedings of the 61st Annual Meeting of the Association for Computational Linguistics (Volume 1: Long Papers), {ACL} 2023, Toronto, Canada, July 9-14, 2023}, pages 12286--12312. Association for Computational Linguistics.

\bibitem[{Longpre et~al.(2021)Longpre, Perisetla, Chen, Ramesh, DuBois, and Singh}]{LongprePCRD021}
Shayne Longpre, Kartik Perisetla, Anthony Chen, Nikhil Ramesh, Chris DuBois, and Sameer Singh. 2021.
\newblock Entity-based knowledge conflicts in question answering.
\newblock In \emph{Proceedings of the 2021 Conference on Empirical Methods in Natural Language Processing, {EMNLP} 2021, Virtual Event / Punta Cana, Dominican Republic, 7-11 November, 2021}, pages 7052--7063. Association for Computational Linguistics.

\bibitem[{Men et~al.(2024)Men, Xu, Zhang, Wang, Lin, Lu, Han, and Chen}]{bi_values}
Xin Men, Mingyu Xu, Qingyu Zhang, Bingning Wang, Hongyu Lin, Yaojie Lu, Xianpei Han, and Weipeng Chen. 2024.
\newblock Shortgpt: Layers in large language models are more redundant than you expect.
\newblock \emph{CoRR}, abs/2403.03853.

\bibitem[{Meng et~al.(2022{\natexlab{a}})Meng, Bau, Andonian, and Belinkov}]{MengBAB22}
Kevin Meng, David Bau, Alex Andonian, and Yonatan Belinkov. 2022{\natexlab{a}}.
\newblock Locating and editing factual associations in {GPT}.
\newblock In \emph{Advances in Neural Information Processing Systems 35: Annual Conference on Neural Information Processing Systems 2022, NeurIPS 2022, New Orleans, LA, USA, November 28 - December 9, 2022}.

\bibitem[{Meng et~al.(2022{\natexlab{b}})Meng, Bau, Andonian, and Belinkov}]{mlp}
Kevin Meng, David Bau, Alex Andonian, and Yonatan Belinkov. 2022{\natexlab{b}}.
\newblock Locating and editing factual associations in {GPT}.
\newblock In \emph{Advances in Neural Information Processing Systems 35: Annual Conference on Neural Information Processing Systems 2022, NeurIPS 2022, New Orleans, LA, USA, November 28 - December 9, 2022}.

\bibitem[{nostalgebraist(2020)}]{nostalgebraist2020interpreting}
nostalgebraist. 2020.
\newblock Interpreting {GPT}: the logit lens.
\newblock \emph{AI Alignment Forum}.

\bibitem[{Olsson et~al.(2022)Olsson, Elhage, Nanda, Joseph, DasSarma, Henighan, Mann, Askell, Bai, Chen et~al.}]{induction-head}
Catherine Olsson, Nelson Elhage, Neel Nanda, Nicholas Joseph, Nova DasSarma, Tom Henighan, Ben Mann, Amanda Askell, Yuntao Bai, Anna Chen, et~al. 2022.
\newblock In-context learning and induction heads.
\newblock \emph{arXiv preprint arXiv:2209.11895}.

\bibitem[{Qiu et~al.(2024)Qiu, Ou, Wu, Li, Liu, and King}]{CLeHE}
Zexuan Qiu, Zijing Ou, Bin Wu, Jingjing Li, Aiwei Liu, and Irwin King. 2024.
\newblock Entropy-based decoding for retrieval-augmented large language models.
\newblock \emph{CoRR}, abs/2406.17519.

\bibitem[{Rajpurkar et~al.(2016)Rajpurkar, Zhang, Lopyrev, and Liang}]{squad}
Pranav Rajpurkar, Jian Zhang, Konstantin Lopyrev, and Percy Liang. 2016.
\newblock Squad: 100, 000+ questions for machine comprehension of text.
\newblock In \emph{Proceedings of the 2016 Conference on Empirical Methods in Natural Language Processing, {EMNLP} 2016, Austin, Texas, USA, November 1-4, 2016}, pages 2383--2392. The Association for Computational Linguistics.

\bibitem[{Ram et~al.(2023)Ram, Levine, Dalmedigos, Muhlgay, Shashua, Leyton-Brown, and Shoham}]{ram-etal-2023-context}
Ori Ram, Yoav Levine, Itay Dalmedigos, Dor Muhlgay, Amnon Shashua, Kevin Leyton-Brown, and Yoav Shoham. 2023.
\newblock \href {https://doi.org/10.1162/tacl_a_00605} {In-context retrieval-augmented language models}.
\newblock \emph{Transactions of the Association for Computational Linguistics}, 11:1316--1331.

\bibitem[{Ren et~al.(2025)Ren, Wang, Qu, Zhao, Liu, Wu, Wen, and Wang}]{RenWQZ00W025}
Ruiyang Ren, Yuhao Wang, Yingqi Qu, Wayne~Xin Zhao, Jing Liu, Hua Wu, Ji{-}Rong Wen, and Haifeng Wang. 2025.
\newblock Investigating the factual knowledge boundary of large language models with retrieval augmentation.
\newblock In \emph{Proceedings of the 31st International Conference on Computational Linguistics, {COLING} 2025, Abu Dhabi, UAE, January 19-24, 2025}, pages 3697--3715. Association for Computational Linguistics.

\bibitem[{Shi et~al.(2024{\natexlab{a}})Shi, Jin, Shen, Dong, Wu, and Xiong}]{IRCAN}
Dan Shi, Renren Jin, Tianhao Shen, Weilong Dong, Xinwei Wu, and Deyi Xiong. 2024{\natexlab{a}}.
\newblock {IRCAN:} mitigating knowledge conflicts in {LLM} generation via identifying and reweighting context-aware neurons.
\newblock \emph{CoRR}, abs/2406.18406.

\bibitem[{Shi et~al.(2024{\natexlab{b}})Shi, Han, Lewis, Tsvetkov, Zettlemoyer, and Yih}]{ShiHLTZY24}
Weijia Shi, Xiaochuang Han, Mike Lewis, Yulia Tsvetkov, Luke Zettlemoyer, and Wen{-}tau Yih. 2024{\natexlab{b}}.
\newblock Trusting your evidence: Hallucinate less with context-aware decoding.
\newblock In \emph{Proceedings of the 2024 Conference of the North American Chapter of the Association for Computational Linguistics: Human Language Technologies: Short Papers, {NAACL} 2024, Mexico City, Mexico, June 16-21, 2024}, pages 783--791. Association for Computational Linguistics.

\bibitem[{Skean et~al.(2024)Skean, Arefin, LeCun, and Shwartz-Ziv}]{skean}
Oscar Skean, Md~Rifat Arefin, Yann LeCun, and Ravid Shwartz-Ziv. 2024.
\newblock Does representation matter? exploring intermediate layers in large language models.
\newblock In \emph{NeurIPs Workshop on Machine Learning and Compression}.

\bibitem[{Sun et~al.(2024)Sun, Zang, Zheng, Song, Xu, Zhang, Yu, and Li}]{REDEEP}
Zhongxiang Sun, Xiaoxue Zang, Kai Zheng, Yang Song, Jun Xu, Xiao Zhang, Weijie Yu, and Han Li. 2024.
\newblock Redeep: Detecting hallucination in retrieval-augmented generation via mechanistic interpretability.
\newblock \emph{CoRR}, abs/2410.11414.

\bibitem[{Xie et~al.(2024)Xie, Zhang, Chen, Lou, and Su}]{Xie0CL024}
Jian Xie, Kai Zhang, Jiangjie Chen, Renze Lou, and Yu~Su. 2024.
\newblock Adaptive chameleon or stubborn sloth: Revealing the behavior of large language models in knowledge conflicts.
\newblock In \emph{The Twelfth International Conference on Learning Representations, {ICLR} 2024, Vienna, Austria, May 7-11, 2024}. OpenReview.net.

\bibitem[{Xin et~al.(2021)Xin, Tang, Yu, and Lin}]{DBLP:conf/eacl/XinTYL21}
Ji~Xin, Raphael Tang, Yaoliang Yu, and Jimmy Lin. 2021.
\newblock Berxit: Early exiting for {BERT} with better fine-tuning and extension to regression.
\newblock In \emph{Proceedings of the 16th Conference of the European Chapter of the Association for Computational Linguistics: Main Volume, {EACL} 2021, Online, April 19 - 23, 2021}, pages 91--104. Association for Computational Linguistics.

\bibitem[{Xu et~al.(2020)Xu, Zhao, Song, Stewart, and Ermon}]{XuZSSE20}
Yilun Xu, Shengjia Zhao, Jiaming Song, Russell Stewart, and Stefano Ermon. 2020.
\newblock A theory of usable information under computational constraints.
\newblock In \emph{8th International Conference on Learning Representations, {ICLR} 2020, Addis Ababa, Ethiopia, April 26-30, 2020}. OpenReview.net.

\bibitem[{Yu et~al.(2023)Yu, Merullo, and Pavlick}]{YuMP23}
Qinan Yu, Jack Merullo, and Ellie Pavlick. 2023.
\newblock Characterizing mechanisms for factual recall in language models.
\newblock In \emph{Proceedings of the 2023 Conference on Empirical Methods in Natural Language Processing, {EMNLP} 2023, Singapore, December 6-10, 2023}, pages 9924--9959. Association for Computational Linguistics.

\bibitem[{Yuan et~al.(2024)Yuan, Yang, Wang, Liu, Zhao, and Liu}]{YuanYWLZL24}
Xiaowei Yuan, Zhao Yang, Yequan Wang, Shengping Liu, Jun Zhao, and Kang Liu. 2024.
\newblock Discerning and resolving knowledge conflicts through adaptive decoding with contextual information-entropy constraint.
\newblock In \emph{Findings of the Association for Computational Linguistics, {ACL} 2024, Bangkok, Thailand and virtual meeting, August 11-16, 2024}, pages 3903--3922. Association for Computational Linguistics.

\bibitem[{Zhao et~al.(2024)Zhao, Zhou, Li, Tang, Wang, Hou, Min, Zhang, Zhang, Dong, Du, Yang, Chen, Chen, Jiang, Ren, Li, Tang, Liu, Liu, Nie, and Wen}]{zhao2024}
Wayne~Xin Zhao, Kun Zhou, Junyi Li, Tianyi Tang, Xiaolei Wang, Yupeng Hou, Yingqian Min, Beichen Zhang, Junjie Zhang, Zican Dong, Yifan Du, Chen Yang, Yushuo Chen, Zhipeng Chen, Jinhao Jiang, Ruiyang Ren, Yifan Li, Xinyu Tang, Zikang Liu, Peiyu Liu, Jian-Yun Nie, and Ji-Rong Wen. 2024.
\newblock \href {https://arxiv.org/abs/2303.18223} {A survey of large language models}.
\newblock \emph{Preprint}, arXiv:2303.18223.

\bibitem[{Zhou et~al.(2024)Zhou, Feng, Zhu, Qian, and Mao}]{nips/ZhouFZQM24}
Hanzhang Zhou, Zijian Feng, Zixiao Zhu, Junlang Qian, and Kezhi Mao. 2024.
\newblock Unibias: Unveiling and mitigating {LLM} bias through internal attention and {FFN} manipulation.
\newblock In \emph{Advances in Neural Information Processing Systems 38: Annual Conference on Neural Information Processing Systems 2024, NeurIPS 2024, Vancouver, BC, Canada, December 10 - 15, 2024}.

\bibitem[{Zhou et~al.(2023)Zhou, Zhang, Poon, and Chen}]{conf/emnlp/ZhouZPC23}
Wenxuan Zhou, Sheng Zhang, Hoifung Poon, and Muhao Chen. 2023.
\newblock Context-faithful prompting for large language models.
\newblock In \emph{Findings of the Association for Computational Linguistics: {EMNLP} 2023, Singapore, December 6-10, 2023}, pages 14544--14556. Association for Computational Linguistics.

\end{thebibliography}

\appendix

\clearpage
\section{Experimental Settings for CounterFact}
\subsection{Data Format in Section 2}\label{appendix:sec2_exp}
We use question $q$ with context $c$ from COUNTERFACT~\cite{MengBAB22}. For a given model, we input the sample data for which the model predicts the correct answer (e.g., "Danielle Darrieux, a native French. The mother tongue of Danielle Darrieux is \_"). In this section, we refer to the token predicted by the model for a given input as the answer.

\begin{tcolorbox}

$c$: \code{\{\{paraphrased prompt\} \{target true\}.\}}

$q$: \code{\{\{prompt\}\}}
\end{tcolorbox}

\subsection{Data Format in Section 3} \label{appendix:se3_exp}
Similarly, we construct the data for computing these two KL divergences based on the CounterFact dataset:

\begin{align*}
    KL_q(l) &= \text{KL}\left[ P(\boldsymbol{h}_l \mid q) \,\|\, P(\boldsymbol{h}_l) \right] \\
    KL_c(l) &= \text{KL}\left[ P(\boldsymbol{h}_l \mid q, c) \,\|\, P(\boldsymbol{h}_l \mid q) \right]
\end{align*}

The term without \( c \) (i.e., \( P(\boldsymbol{h}_l \mid q) \)) is generated by excluding \( c \) from the input. 
\begin{tcolorbox}
$q$: \code{\{\{prompt\}\}}
\end{tcolorbox}
The term without both \( q \) and \( c \) (i.e., \( P(\boldsymbol{h}_l) \)) is derived by providing an empty input.
\begin{tcolorbox}
\code{\{$\emptyset$\}}
\end{tcolorbox}

\section{Logit Lens} \label{appendix:logitlens}

The \(\operatorname{LogitLens}\) is a technique that decodes hidden states \(\boldsymbol{h}^{l}\) directly into the vocabulary distribution using the LayerNorm and the unembedding matrix of the LLM for interpretability~\cite{nostalgebraist2020interpreting}:

\begin{equation}
    \operatorname{LogitLens}\left(\boldsymbol{h}^{l}\right) = \operatorname{LayerNorm}(\boldsymbol{h}^{l}) \boldsymbol{W}_U
\end{equation}
The final layer's residual stream state is then projected into the vocabulary space using the \textit{Unembedding Matrix} \(\boldsymbol{W}_U \in \mathbb{R}^{d \times |\mathcal{V}|}\) and normalized via the softmax function to produce a probability distribution over the vocabulary, from which a new token is sampled.

This approach has been validated in various studies as an effective method for interpreting LLMs' weight matrices or hidden states~\cite{YuMP23,nips/0001LV23,nips/ZhouFZQM24}.

\section{Residual Stream} \label{appendix:residual}

We interpret transformer decoder-only architecture (also known as GPT-like) through the perspective of the residual stream \citep{elhage2021mathematical, induction-head}. Due to the residual connections in Transformers, each layer \( l \) takes a hidden state \(\boldsymbol{h}^{l-1}\) as input and adds information obtained from its \textit{Attention Heads} and \textit{Feed-Forward Networks} (FFNs) to the hidden state via the residual connection. In this context, the hidden state acts as a residual stream passed through the layers, with each attention and FFN contributing to the final prediction by adding information to the residual stream, resulting in the \textit{Residual Stream States}. 
Formally, the hidden state \(\boldsymbol{h}^{l}\) at layer $l$ is calculated as:
\begin{align*}
    \boldsymbol{h}^{l} &=\boldsymbol{h}^{l-1} +\text{MHA}(\boldsymbol{h}^{l-1}) + \text{FFN}(\boldsymbol{a}^{l}) \nonumber \\
    &= \boldsymbol{h}^{l-1} + \boldsymbol{a}^{l} + \boldsymbol{m}^{l}
\end{align*}
where $\boldsymbol{a}^{l}$ and $\boldsymbol{m}^{l}$ are the outputs from the MHA and FFN block in the $l$-th layer. Both quantities are dependent on $\boldsymbol{h}^{l-1}$ and can thus be formulated as functions of it (See Eq.~\ref{eq:s_i}).

Then, the hidden state \(\boldsymbol{h}^{l+1}\) at layer $l+1$ is calculated as:
\begin{align*}
    \boldsymbol{h}^{l+1} &= \boldsymbol{h}^{l} + \boldsymbol{a}^{l+1} + \boldsymbol{m}^{l+1} \nonumber\\
    & = \boldsymbol{h}^{l-1} + \boldsymbol{a}^{l} + \boldsymbol{m}^{l} + \boldsymbol{a}^{l+1} + \boldsymbol{m}^{l+1} \nonumber\\
    & = \boldsymbol{h}^{l-1} + \sum_{k=l}^{l+1} \boldsymbol{a}^{k} +\sum_{k=l}^{l+1} \boldsymbol{m}^{k} 
\end{align*}

Consequently, the hidden state \(\boldsymbol{h}^{N}_i\) at the final layer $N (N\geq l)$ can be calculated as:
\begin{align*} 
    \boldsymbol{h}^{N} &= \boldsymbol{h}^{N-1} + \boldsymbol{a}^{N} + \boldsymbol{m}^{N} \nonumber\\
    & = \boldsymbol{h}^{N-2} + \boldsymbol{a}^{N-1} + \boldsymbol{m}^{N-1} + \boldsymbol{a}^{N} + \boldsymbol{m}^{N} \nonumber\\
    & = ... \nonumber \\
    & = \boldsymbol{h}^{l} + \sum_{k=l+1}^N \boldsymbol{a}^{k} +\sum_{k=l+1}^N \boldsymbol{m}^{k} 
\end{align*}
where $\mathbf{s}_i$ represents the sum of the contributions from the subsequent layers to the final layer.

The final layer's residual stream state is then projected into the vocabulary space using the \textit{Unembedding Matrix} \(\boldsymbol{W}_U \in \mathbb{R}^{d \times |\mathcal{V}|}\). The final output logits of the LLM can be expressed as:
\begin{align}\label{eq:s_i}
      \text{logits}_{N} &= \boldsymbol{h}^{l} \boldsymbol{W}_U + \left(\sum_{k=l+1}^N \boldsymbol{a}^{k} +\sum_{k=l+1}^N \boldsymbol{m}^{k} \right)\boldsymbol{W}_U \nonumber \\
      & = \mathbf{v} + \mathbf{u}(\mathbf{v})
\end{align}
For analytical simplicity, we ignore the final LayerNorm function following~\citet{elhage2021mathematical} and ~\citet{REDEEP}. It adds a fair amount of complexity to consider explicitly, and up to a variable scaling, layer norm can be merged into adjacent weights. Conceptually, $\mathbf{v}$ and $\mathbf{u}(\mathbf{v})$ capture the information encoded in layer $l$ and later layer $>l$ respectively.

Finally, the logit would be normalized via the softmax function to produce a probability distribution over the vocabulary, from which a new token is sampled.
\begin{equation} \label{eq:ori_p}
    \text{P} = \frac{e^{ v_k + u_k(v_k)}}{\sum_{j\in {|Vocab|}} e^{ v_j + u_j(v_j)}}
\end{equation}

\section{Theoretical Support for Enhancement}\label{appendix:enhancement}
\subsection{Analysis on the Amplification}\label{appendix:amplification}
Suppose we amplify the hidden states at layer $ l $ by the factor $\alpha$:
$$\boldsymbol{h}^{l}_{\text{modified}} = \alpha \boldsymbol{h}^{l}$$
We will analyze how this scaling affects the subsequent computations. Here Let's take LayerNorm as an example.

\paragraph{At Layer $ l+1$.}
Since LayerNorm is scale-invariant:
\begin{align}
    \tilde{\boldsymbol{h}}^{l}_{\text{modified}} &= \text{LayerNorm}(\alpha \boldsymbol{h}^{l})  \nonumber\\
    &= \text{LayerNorm}(\boldsymbol{h}^{l})
    = \tilde{\boldsymbol{h}}^{l} \nonumber
\end{align}
The scaling by $ \alpha $ has no effect on the output of the first LayerNorm. Therefore, the input to the MHA sublayer remains unchanged:
$$\boldsymbol{a}^{l+1} = \text{MHA}(\tilde{\boldsymbol{h}}^{l})$$
After the MHA sublayer, the residual connection adds $ \boldsymbol{a}^{l} $ back to the scaled $ \boldsymbol{h}^{l} $:
$$\boldsymbol{h}^{l+1}_{\text{modified}} = \boldsymbol{h}^{l}_{\text{modified}} + \boldsymbol{a}^{l+1} = \alpha \boldsymbol{h}^{l} + \boldsymbol{a}^{l+1}$$
Then the scaled hidden state $ \alpha h^{l} $ is now part of $ \boldsymbol{h}^{l+1} $, which will be normalized for FFN input.
\begin{align*}\label{eq:norm}
    \tilde{\boldsymbol{h}}^{l+1}_{\text{modified}} &= \text{LayerNorm}(\boldsymbol{h}^{l+1}_{\text{modified}}) \nonumber \\
    & = \gamma \odot \frac{\alpha \boldsymbol{h}^{l} + \boldsymbol{a}^{l} - \mu_{\text{modified}}^{l+1}}{\sigma_{\text{modified}}^{l+1}} + \delta 
\end{align*}
where 
\begin{align}
    \mu_{\text{modified}}^{l+1} &= \frac{1}{D} \sum_{i=1}^D \left( \alpha \boldsymbol{h}^{l} + \boldsymbol{a}^{l} \right)= \alpha \mu^{l} + \mu_{a} \nonumber \\
    \sigma_{\text{modified}}^{l+1} &= \sqrt{\frac{1}{D} \sum_{i=1}^D \left( \alpha \boldsymbol{h}^{l} + \boldsymbol{a}^{l} - \mu_{\text{modified}}^{l+1} \right)^2} \nonumber
\end{align}
The normalization does not cancel out $\alpha$ completely because $\boldsymbol{a}^{l}$ is not scaled. Therefore, the information in $\boldsymbol{a}$ is relatively compressed. Then,
$$\boldsymbol{m}^{l+1}_{\text{modified}} = \text{MLP}(\tilde{\boldsymbol{h}}^{l+1}_{\text{modified}})$$
The output at layer $l+1$ is:
\begin{align}
    \boldsymbol{h}^{l+1}_{\text{modified}} &= \boldsymbol{h}^{l}_{\text{modified}} + \boldsymbol{a}^{l+1}+ \boldsymbol{m}^{l+1}_{\text{modified}}\nonumber\\
    &= \alpha \boldsymbol{h}^{l} + \boldsymbol{a}^{l+1} + \boldsymbol{m}^{l+1}_{\text{modified}} \nonumber
\end{align}

\paragraph{At Layer $k > l+1$.}
The altered hidden state $ h^{l+1}_{\text{modified}} $ affects all subsequent layers in a similar fashion. And the amplification effect of $\alpha$ persists due to the residual connections.

The hidden state can be represented recursively:
\begin{align}
    \boldsymbol{h}^{k}_{\text{modified}} = \alpha \boldsymbol{h}^{l} &+ \boldsymbol{a}^{l+1} + \boldsymbol{m}^{k+1}_{\text{modified}} \nonumber\\
    &+\sum_{i=l+2}^{k} \left( \boldsymbol{a}^{i}_{\text{modified}} + \boldsymbol{m}^{i}_{\text{modified}} \right) \nonumber
\end{align}

Therefore, the amplification effect of $\alpha $ \textbf{accumulates} through the residual connections and affects all subsequent layers.

\paragraph{At Final Layer $N$.}
The logits are computed using the final hidden state $ \boldsymbol{h}^{N}_{\text{modified}} $:
$$\text{logits}_{\text{modified}} = \boldsymbol{h}^{N}_{\text{modified}} \boldsymbol{W}_U$$
Breaking Down $ \boldsymbol{h}^{N}_{\text{modified}} $, the logits can be calculated as:
\begin{align}
    \text{logits}_{\text{modified}} &= \alpha \boldsymbol{h}^{l} \boldsymbol{W}_U + (\boldsymbol{a}^{l+1} + \boldsymbol{m}^{l+1}_{\text{modified}} \nonumber \\
    & +\sum_{i=l+2}^{N} ( \boldsymbol{a}^{i}_{\text{modified}} + \boldsymbol{m}^{i}_{\text{modified}} ))\boldsymbol{W}_U \nonumber
\end{align}
Finally, the softmax probabilities become:
\begin{equation}\label{eq:p_modified1}
    \text{P}_{\text{modified}} = \frac{e^{\alpha v_k + u^{(A)}_k(\alpha v)}}{\sum_{j\in {|Vocab|}} e^{\alpha v_j + u^{(A)}_j(\alpha v)}} 
\end{equation}
where:
$ \boldsymbol{v} = \boldsymbol{h}^{L} \boldsymbol{W}_U $, and $ \boldsymbol{u}^{(A)}(\cdot) $ includes contributions from the subsequent layers by amplification.

This leads to a proportional change in the logits and alters the softmax probability distribution, potentially affecting the model's predictions.

In correspondence with Formula~\ref{eq:ori_p}, the $\mathcal V$-entropy is derived as follows:
\begin{equation*}
    H_{\mathcal V_f}(\alpha) \defeq \mathbb{E}[\ln \sum_j e^{\alpha v_j + u_j(\alpha v_j)} - \left(\alpha v_k + u_k(\alpha v_k)\right)] \nonumber
\end{equation*}
\subsection{Analysis on the Additional Residual Connection} \label{appendix:residual connextion}
Suppose we introduce an additional residual connection at layer \( l \), which directly propagates \( \boldsymbol{h}^{l} \) to $\alpha$ subsequent layers:

\paragraph{At Layer \( l+1 \).}  
With the inclusion of the new residual connection, the hidden state is modified as follows:  
\begin{align*}
    \boldsymbol{h}^{l+1}_{\text{modified}} &=( \boldsymbol{h}^{l} + \boldsymbol{a}^{l+1} + \boldsymbol{m}^{l+1}) + \boldsymbol{h}^{l} \nonumber \\
    &= 2\boldsymbol{h}^{l} + \boldsymbol{a}^{l+1} + \boldsymbol{m}^{l+1}
\end{align*}
\paragraph{At Layer $k$ where $l+1 < k \leq l+\alpha$.}
Due to the cumulative effect of the residual connections, the hidden state at layer $k$ can be expressed as:
\begin{align}
\boldsymbol{h}^{k}_{\text{modified}} &= (k-l+1)\boldsymbol{h}^{l} + \boldsymbol{a}^{l+1} + \boldsymbol{m}^{l+1} \nonumber\\
&+\sum_{i=l+2}^{k} (\boldsymbol{a}^{i}_{\text{modified}} + \boldsymbol{m}^{i}_{\text{modified}}) \nonumber
\end{align}
where $(k-l+1)$ represents the number of times the residual connection has been accumulated.
\paragraph{At the Final Layer $N$.}
The final logits are computed as:
\begin{align}
&\text{logits}_{\text{modified}} = \boldsymbol{h}^{N}_{\text{modified}} \boldsymbol{W}_U \nonumber\\
&= (\alpha+1)\boldsymbol{h}^{l}\boldsymbol{W}_U + (\boldsymbol{a}^{l+1} + \boldsymbol{m}^{l+1} \nonumber\\
&+\sum_{i=l+2}^{N} (\boldsymbol{a}^{i}_{\text{modified}} + \boldsymbol{m}^{i}_{\text{modified}}))\boldsymbol{W}_U \nonumber
\end{align}
Similar to the previous deduction (Appendix~\ref{appendix:amplification}), the final softmax probability distribution is given by:
\begin{equation}\label{eq:p_modified2}
\text{P}_{\text{modified}} = \frac{e^{(\alpha+1) v_k + u^{(R)}_k((\alpha+1) v)}}{\sum_{j\in {|Vocab|}} e^{(\alpha+1) v_j + u^{(R)}_j((\alpha+1) v)}}
\end{equation}
where $\boldsymbol{v} = \boldsymbol{h}^{(l)} \boldsymbol{W}_U$. While $\boldsymbol{u}^{(R)}(\cdot)$ encapsulates the contributions from the subsequent layers, it differs fundamentally from the $\boldsymbol{u}^{(A)}(\cdot)$ function in the Amplification method, as the underlying modifications to the information flow in these two approaches are inherently distinct.

\paragraph{Compared with Amplification.} This analysis reveals that adding cumulative residual connections provides a structured approach to amplifying the influence of intermediate layer representations. While there are subtle differences in how these methods affect subsequent layers, we empirically compare their performances through experiments in Section~\ref{exp:experiments}.

\section{Proof for Proposition 3.1} \label{appendix:proof}
\paragraph{Proposition.}
Let $\alpha$ denote the scaling factor applied to the hidden states at this layer. 
If $k=\arg\max_j v_j$, then 
\begin{equation*}
    \lim_{\alpha \to \infty} H_{\mathcal V_f}(\alpha) \approx 0
\end{equation*}

\begin{proof}
First, consider the following decomposition:
\begin{align}
    \sum_j &e^{\alpha v_j + u_j(\alpha v)} = e^{\alpha v_k + u_k(\alpha v)}\cdot  \nonumber\\
    &\left(1 + \sum_{j \neq k} e^{\alpha (v_j - v_k) + u_j(\alpha v_j) - u_k(\alpha v_k)}\right) \nonumber
\end{align}
When $j \neq k$, $v_j - v_k < 0$. All terms with $j \neq k$ exponentially diminish as $\alpha$ increases\footnote{Since $\alpha$ approaches zero, $u_j(\alpha v) - u_k(\alpha v)$ represents the difference between two extremely small quantities. It can be negligible compared to  the $\alpha (v_j - v_k)$.}. Therefore:
\begin{equation}
\lim_{\alpha \to \infty}\sum_j e^{\alpha v_j + u_j(\alpha v_j)} \approx e^{\alpha v_k + u_k(\alpha v_k)} \left(1 + \epsilon(\alpha)\right) \nonumber
\end{equation}
where $\epsilon(\alpha) \to 0$ as $\alpha \to \infty$.

Applied to the Formula~\ref{eq:h_alpha}, we have
\begin{align*}
\lim_{\alpha \to \infty}H_{\mathcal V_f}(\alpha) &\approx \mathbb{E}\Big[\alpha v_k + u_k(\alpha v)+ \ln \left(1 + \epsilon(\alpha)\right) \nonumber\\
&\qquad\qquad - (\alpha v_k + u_k(\alpha v))\Big] \nonumber\\
&= \mathbb{E}\left[\ln \left(1 + \epsilon(\alpha)\right)\right] \nonumber\\
&= 0
\end{align*}
Therefore, the result is proven.
\end{proof}

\section{Dataset Details} \label{appendix:data_detail}
\paragraph{CounterFact.} The CounterFact~\cite{MengBAB22} dataset is derived from the PARAREL dataset~\cite{ElazarKRRHSG21} and contains knowledge tuples of the kind $t^c = (s, r, o^c)$, where $s$ is the subject, $r$ is the relation and $o$ is the object. These tuples are constructed using entities listed in Wikidata. The data are accompanied by handwritten paraphrased prompts for each sample. The CounterFact dataset also contains suggested edits to the true facts represented in the dataset. For this study, the set of counterfactual edits are not used. 

\paragraph{NaturalQuestions.} NQ~\cite{NQ_KwiatkowskiPRCP19} consists of real-world information-seeking queries issued to the Google search engine and their corresponding long answers (gold evidence passage) and short answers (one or more entities). In our study, we employ the long answers as the input context and short answers as the ground truth, and conduct evaluations on the dev set.

\paragraph{NQ-Swap} NQ-Swap is based on the NQ dataset, where the objective is to answer questions based on a reliable gold document. To generate NQ-Swap, \citet{LongprePCRD021} first identify questions in NQ with named entity answers, find the supportive document for each question and then replace the gold answer entity in the document with a random entity. A faithful LM should generate the replaced entity as the answer when given the question and modified document.

\paragraph{SQuAD.} The SQuAD~\cite{squad} 1.1 is a common QA benchmark. It includes questions posed by human annotators on a given Wikipedia paragraph, where the answer to each question is a segment of text (or span) from the paragraph. In our experiments, we conduct experiments on the dev for evaluation. 

\paragraph{StrategyQA.} StrategyQA~\cite{strategyqa} is a fact reasoning benchmark that necessitates the implicit question decomposition into reasoning steps. Built around Wikipedia terms, these questions are accompanied by multiple evidence paragraphs.  The model is expected to provide a True or False answer. We concatenate question-relevant evidences to form the input context. We adopt the training set for evaluation, considering the volume of data.

\subsection{Posteriori judgement for CounterFact} \label{appendix:prior}
We delineates the process of identifying knowledge boundary of "unknown" and "known" contexts. The evaluation is based on the accuracy of the model's responses when context is not provided. The scenarios are divided into two categories:
\begin{itemize}
\item \textbf{Unkown}: This category refers to instances where the model is unable to provide the correct answer without relying on the provided context. Such cases indicate that the external contextual knowledge represents information not contained within the model's inherent parametric knowledge.
\item \textbf{Known}: This category describes scenarios in which the model can accurately answer a question without requiring its corresponding context. These instances demonstrate that the model has internalized the relevant knowledge, reflecting an alignment between its parametric knowledge and the external contextual information.
\end{itemize}

\section{Decoding Strategies} \label{appendix:decoding}
\paragraph{Contrastive Decoding (CD)} In our experiments, we employ the distribution $g(y_t)$ with a certain threshold as a baseline decoding method, referred to as the CD~\cite{LiHFLEHZL23} method. We modify the original object of CD (computes the distribution discrepancy between an small amateur model and an expert larger model) to simulate the form of $g(y_t)$.  
\begin{align*}
CD &= \log p(y_t|x,y<t)-p(y_t|y<t) \\
&= \log g(y_t) 
\end{align*}
The threshold is same as in the original CD method:
\begin{align*}
&\mathcal{V}_{\text{head}}(y_{<t}) = \\
&\left\{ y_t \in \mathcal{V} : p(y_t | y_{<t}) \geq 0.1 \cdot \max_{y} p(y | y_{<t}) \right\} \nonumber
\end{align*}
Here, we represent the input context as $x$. CD adopts the object of difference between the output likelihood when inputs are presented with and without input context. It enhances the influence of the context for high-probability words within a crude threshold.

\paragraph{Context-Aware Decoding (CAD)}
In CAD~\cite{ShiHLTZY24} method, the output probability is a product-of-experts of the original output probability and PMI weighted by $\alpha=0.5$ as follow:
\begin{align*}
    &y_t \sim \operatorname{softmax}[(1+\alpha) \operatorname{logit}_\theta(y_t \mid \boldsymbol{c}, \boldsymbol{x}, \boldsymbol{y}_{<t})\\
    &\qquad \qquad - \alpha \operatorname{logit}_\theta(y_t \mid \boldsymbol{x}, \boldsymbol{y}_{<t})]
\end{align*} 

\paragraph{COntextual Information-Entropy Constraint Decoding (COIECD)}
First, the contextual contrastive object $g$ is calculated to quantify the divergence between $p_1$ and $p_2$:
\begin{equation*}
    g(y_t) = \log p_2(y_t) - \log p_1(y_t)
\end{equation*}
where
\begin{align*}
    p_{1}(y_t) &= p(y_t|\boldsymbol{x},\boldsymbol{y}_{<t}) \\
    p_{2}(y_t) &= p(y_t|\boldsymbol{x},\boldsymbol{c},\boldsymbol{y}_{<t})
\end{align*}
The $g$ is to factor out the model's inherent memory and favor the contextual knowledge. 

The contextual information-entropy constraint is utilized with $g$ on the output distribution $\pi$ as:
\begin{align}
&\quad \log\pi(y_t \mid\boldsymbol{x},\boldsymbol{c},\boldsymbol{y}_{<t}) \\
&=
\begin{cases}
\log p_1(y_t) +  \alpha \cdot  g(y_t) & \text{if } y_t \in \mathcal{C}(\boldsymbol{y}_{<t}), \\
\log p_2(y_t) +  \alpha \cdot g(y_t) & \text{otherwise} .
\end{cases}\nonumber
\end{align}
where $\alpha$ is a scaling weight to control the contextual impact.
The final decoding strategy can be formalized as:
\begin{equation}
    y_t \sim \mathrm{softmax}[\log\pi(y_t \mid\boldsymbol{x},\boldsymbol{c},\boldsymbol{y}_{<t})] 
\end{equation}
In this way, COIECD strikes a balance between the two sources of knowledge to achieve a more effective and holistic decoding strategy.

\section{Layer Selection by CaLE}
Here, we present some models that employ the CaLE method, as shown in Tables~\ref{tab:main-results} and \ref{tab:nq}, which enhance various layers selected through both supervised and unsupervised identification, as indicated in Table~\ref{tab:layer}. Our findings reveal that nearly all of the selected layers are distributed in the middle to later stages, suggesting that intervening at deeper layers is a more effective choice.

\begin{table}[htbp]
\centering
\tiny
\begin{tabular}{l|cc|cc|cc}
\midrule
Layer selected& \multicolumn{2}{c|}{\textbf{CounterFact}} & \multicolumn{2}{c|}{\textbf{NQ}} & \multicolumn{2}{c}{\textbf{NQ-swap}} \\
by CaLE& sup. & unsup. & sup. & unsup. & sup. & unsup. \\
\midrule
Llama2-7B & 25 & 25 & 26 & 26 & 15 & 26 \\
Llama3.1-8B & 23 & 26 & 28 & 25 & 19 & 23 \\
Llama3.2-3B & 23 & 23 & - & - & - & - \\
Mistral-7B & - & - & 25 & 25 & 15 & 25 \\
Gemma2-9B & - & - & 32 & 39 & 29 & 25 \\
\midrule
& \multicolumn{2}{c|}{\textbf{SQuAD}} & \multicolumn{2}{c|}{\textbf{StrategyQA}} & & \\
& sup. & unsup. & sup. & unsup. & & \\
\midrule
Llama2-7B & 22 & 26 & 22 & 21 & & \\
Llama3.1-8B & 26 & 26 & 30 & 25 & & \\
Mistral-7B & 25 & 27 & 22 & 30 & & \\
Gemma2-9B & 36 & 38 & 35 & 36 & & \\
\bottomrule
\end{tabular}
\caption{Layer selection of CaLE across different datasets and models. The $\mathbf{unsup.}$ and $\mathbf{sup.}$ denote the unsupervised and supervised CaLE methods. }
\label{tab:layer}
\end{table}
\end{document}